\documentclass{article} 
\usepackage{booktabs}
\usepackage{iclr2024,times}

\usepackage{amsmath,amsfonts,bm}

\def\eqref#1{equation~\ref{#1}}

\def\1{\bm{1}}

\def\va{{\bm{a}}}

\def\vc{{\bm{c}}}
\def\vd{{\bm{d}}}

\def\vn{{\bm{n}}}
\def\vo{{\bm{o}}}
\def\vp{{\bm{p}}}

\def\vv{{\bm{v}}}
\def\vw{{\bm{w}}}
\def\vx{{\bm{x}}}
\def\vy{{\bm{y}}}

\DeclareMathAlphabet{\mathsfit}{\encodingdefault}{\sfdefault}{m}{sl}
\SetMathAlphabet{\mathsfit}{bold}{\encodingdefault}{\sfdefault}{bx}{n}

\def\sC{{\mathbb{C}}}

\def\sN{{\mathbb{N}}}

\def\sP{{\mathbb{P}}}

\usepackage{subcaption}
\usepackage{amssymb}
\usepackage{hyperref}
\usepackage{url}
\usepackage{graphicx}
\usepackage{amsmath} 
\usepackage[nonumberlist, toc, nopostdot]{glossaries}
\usepackage{enumitem}
\graphicspath{ {./figures/} }
\makeglossaries

\newglossaryentry{ablation}{
    name={ablation},
    description={A technique where we eliminate the contribution of a particular component to a model's output (usually by replacing the component's output with zeros or the mean over some dataset or a random sample from some dataset) in order to demonstrate the magnitude of its importance.}
}
\newglossaryentry{activation addition}{
    name={activation addition},
    description={Formerly called ``activation steering", a technique from \citet{turner2023activation} where a vector is added to the residual stream at a certain position (or all positions) and layer during each forward pass while generating sentence completions. In our case, the vector is the sentiment direction.}
}
\newglossaryentry{activation patching}{
    name={activation patching},
    description={A technique introduced in \cite{rome}, under the name `causal tracing', which uses a causal intervention to identify which activations in a model matter for producing some output. It runs the model on some `clean' input, replaces (patches) an activation with that same activation on `flipped' input, and sees how much that shifts the output from `clean' to `flipped'.},
}
\newglossaryentry{activation steering}{
    name={activation steering},
    description={See \gls{activation addition}},
    see={activation addition},
}
\newglossaryentry{das}{
    name={DAS},
    description={Distributed Alignment Search \citep{das} uses gradient descent to train a rotation matrix representating an orthonormal change of basis to one better aligned with the model's features. We mostly focus on a special case of finding a singular critical direction, where we patch along the first dimension of the rotated basis and then use a smooth \gls{patching metric} (such as the logit difference between positive and negative completions) as the objective to be minimised.}
}
\newglossaryentry{directional activation patching}{
    name={directional activation patching},
    description={A variant of activation patching introduced in this paper where we only patch a single dimension from a counterfactual activation. That is, for prompts $x_\text{orig}$ and $x_\text{new}$, direction $\vd$, a set of model components $\sC$, we run a forward pass on $x_\text{orig}$ but for each component in $\sC$, we patch/replace the output $\vo_\text{orig}$ with $\vo_\text{orig} - \vo_\text{orig} \cdot \vd + \vo_\text{new} \cdot \vd$. This is equivalent to activation patching a single neuron, but done in a rotated basis (where $\vd$ is the first column of the rotation matrix).},
}
\newglossaryentry{directional patching}{
    name={directional patching},
    description={See \gls{directional activation patching}.},
    see={directional activation patching}
}
\newglossaryentry{logit difference}{
    name={logit difference},
    description={The difference between the logits given to a particular pair of completions. To reduce noise, we can generalise this to the \textit{average} difference between two sets of completions. In our case, the dichotomy of completions generally represent positive vs. negative sentiment.},
}
\newglossaryentry{mean ablation}{
    name={mean ablation},
    description={A type of ablation method, where we seek to eliminate the contribution of a particular component to demonstrate its importance, where we replace a particular set of activations with their mean over an appropriate dataset.},
}
\newglossaryentry{ood}{
    name={OOD},
    description={``Out-of-distribution": we use this to mean a stronger criterion than out of sample where the distribution has shifted significantly between train and test.}
}
\newglossaryentry{oos}{
    name={OOS},
    description={``Out-of-sample": when a model or method is evaluated on a distinct test set from that on which it was trained.}
}
\newglossaryentry{patching metric}{
    name={patching metric},
    description={A summary statistic used to quantify the results of an activation patching experiment. By default here we use the percentage change in logit difference as in \cite{ioi}.}
}
\newglossaryentry{path patching}{
    name={path patching},
    description={A variant of activation patching introduced in \cite{ioi} in which only the activations related to the residual stream paths between two sets of endpoints (senders and receivers) are patched, but the remainder of the network upstream of the receivers is frozen. Given a set $R$ of receivers, a sender attention head $h$, and paths $P$ between $h$ and each of $R$, activations from the mirrored dataset are patched into $P$ while keeping the remainder of the network fixed (aside from everything downstream of $R$)},
}
\newglossaryentry{sst}{
    name={SST},
    description={Stanford Sentiment Treebank is a labelled sentiment dataset from \citet{treebank} described in Section \ref{section:datasets}.}
}

\title{Linear Representations of Sentiment\\ in Large Language Models}

\author{
Curt Tigges$^{\text{*}\clubsuit}$, 
Oskar John Hollinsworth$^{\text{*}\heartsuit}$, 
Atticus Geiger$^{\spadesuit \bigstar}$, 
Neel Nanda$^{\diamondsuit}$ \\
$^{\clubsuit}$EleutherAI Institute,
$^{\heartsuit}$SERI MATS,
$^{\spadesuit}$Stanford University,
$^{\bigstar}$Pr(Ai)$^2$R Group,
$^{\diamondsuit}$Independent \\
$^{\text{*}}$Equal primary authors (order random)
}

\newcommand{\ablateDirectionTreebankLDDrop}{71\% }
\newcommand{\ablateDirectionTreebankAccDrop}{38\% }
\newcommand{\ablateDirectionTreebankAcc}{62\% }
\newcommand{\ablateCommasDirectionTreebankLDDrop}{18\% }
\newcommand{\ablateCommasDirectionTreebankAccDrop}{18\% }

\newcommand{\ablateCommasMeanTreebankLDDrop}{17\% }

\newcommand{\ablateCommasMeanPatchTreebankAccDrop}{19\% }

\newcommand{\simpleTrainSize}{30 }

\newcommand{\owtNbins}{20 }
\newcommand{\owtNsamples}{20 }
\newcommand{\owtNcontext}{20 }
\newcommand{\maxLDratio}{15\% }

\newcommand{\tokens}{\fontfamily{qhv}\selectfont \footnotesize}

\iclrfinalcopy 

\begin{document}

\maketitle

\begin{abstract}
Sentiment is a pervasive feature in natural language text, yet it is an open question how sentiment is represented within Large Language Models (LLMs). In this study, we reveal that across a range of models, sentiment is represented linearly: a single direction in activation space mostly captures the feature across a range of tasks with one extreme for positive and the other for negative. Through causal interventions, we isolate this direction and show it is causally relevant in both toy tasks and real world datasets such as Stanford Sentiment Treebank. Through this case study we model a thorough investigation of what a single direction means on a broad data distribution.

We further uncover the mechanisms that involve this direction, highlighting the roles of a small subset of attention heads and neurons. Finally, we discover a phenomenon which we term the summarization motif: sentiment is not solely represented on emotionally charged words, but is additionally summarised at intermediate positions without inherent sentiment, such as punctuation and names. We show that in Stanford Sentiment Treebank zero-shot classification, 76\% of above-chance classification accuracy is lost when ablating the sentiment direction, nearly half of which (36\%) is due to ablating the summarized sentiment direction exclusively at comma positions.

\end{abstract}

\section{Introduction}\label{introduction}
Large language models (LLMs) have displayed increasingly impressive capabilities \citep{brown2020language,radford2019language,bubeck2023sparks}, but their internal workings remain poorly understood. Nevertheless, recent evidence \citep{li2023emergent} has suggested that LLMs are capable of forming models of the world, i.e., inferring hidden variables of the data generation process rather than simply modeling surface word co-occurrence statistics. There is significant interest (\citet{christiano2021eliciting}, \citet{burns2022discovering}) in deciphering the latent structure of such representations.

In this work, we investigate how LLMs represent sentiment, a variable in the data generation process that is relevant and interesting across a wide variety of language tasks \citep{cui2023survey}. Approaching our investigations through the frame of causal mediation analysis \citep{vig2020causal,pearl2022direct, Geiger-etal:2023:CA}, we show that these sentiment features are represented linearly by the models, are causally significant, and are utilized by human-interpretable circuits
\citep{olah2020zoom,elhage2021mathematical}.

We find the existence of a single direction scientifically interesting as further evidence for the linear representation hypothesis \citep{mikolov-etal-2013-linguistic,elhage2022superposition}-- that models tend to extract properties of the input and internally represent them as directions in activation space. Understanding the structure of internal representations is crucial to begin to decode them, and linear representations are particularly amenable to detailed reverse-engineering \citep{nanda2023emergent}.

We show evidence of a phenomenon we have labeled the ``summarization motif", where rather than sentiment being directly moved from valenced tokens to the final token, it is first aggregated on intermediate summarization tokens without inherent valence such as commas, periods and particular nouns.\footnote{Our use of the term ``summarization'' is distinct from typical NLP summarization tasks} This summarization structure for next token prediction can be seen as a naturally emerging analogue to the explicit classification token in BERT-like models \citep{devlin2018bert}. We show that the sentiment stored on summarization tokens is causally relevant for the final prediction. We find this an intriguing example of an ``information bottleneck", where the data generation process is funnelled through a small subset of tokens used as information stores. Understanding the existence and location of information bottlenecks is a key first step to deciphering world models. This finding additionally suggests the models' ability to create summaries at various levels of abstraction, in this case a sentence or clause rather than a token. 


Our contributions are as follows. In Section \ref{section:direction-finding}, we demonstrate methods for finding a \textbf{linear representation of sentiment} using a toy dataset and show that this direction correlates with sentiment information in the wild and matters causally in a crowdsourced dataset. In Section \ref{section:circuits}, we show through \gls{activation patching} \citep{vig2020causal, geiger-etal-2020-neural} and ablations that the learned sentiment direction captures \textbf{summarization behavior} that is causally important to circuits performing sentiment tasks. Through this case study, we model an investigation of what a single interpretable direction means on the full data distribution.

\begin{figure}
    \centering

    \begin{subfigure}{0.24\linewidth}
        \centering
        \includegraphics[width=\linewidth, clip, trim={0 7cm 8cm 0}]{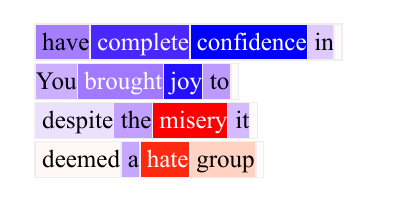}
        \caption{Nouns}
        \label{fig:neuroscope-nouns}
    \end{subfigure}
    \hfill
    \begin{subfigure}{0.24\linewidth}
        \centering
        \includegraphics[width=\linewidth, clip, trim={0 7cm 8cm 0}]{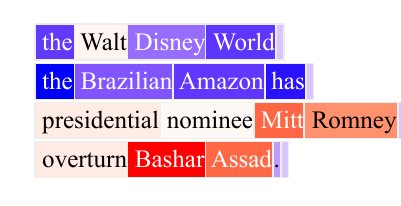}
        \caption{Proper Nouns}
        \label{fig:neuroscope-proper-nouns}
    \end{subfigure}%
    \hfill
    \begin{subfigure}{0.24\linewidth}
        \centering
        \includegraphics[width=\linewidth, clip, trim={0 7cm 8cm 0}]{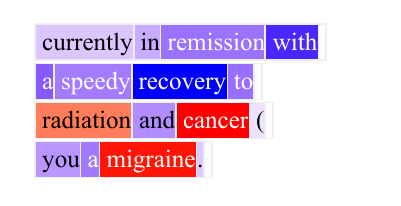}
        \caption{Medical}
        \label{fig:neuroscope-medical}
    \end{subfigure}%
\hfill
    \begin{subfigure}{0.24\linewidth}
        \centering
        \includegraphics[width=\linewidth, clip, trim={0 7cm 8cm 0}]{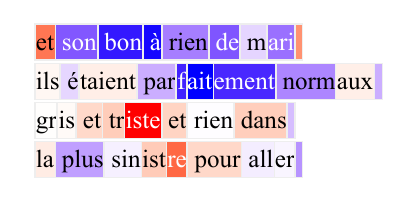}
        \caption{French}
        \label{fig:neuroscope-french}
    \end{subfigure} 
    \caption{Visual verification that a single direction captures sentiment across diverse contexts. Color represents the projection onto this direction, blue is positive and red is negative. Examples (\ref{fig:neuroscope-nouns}-\ref{fig:neuroscope-medical}) show the $K$-means sentiment direction for the first layer of GPT2-small on samples from OpenWebText. Example \ref{fig:neuroscope-french} shows the $K$-means sentiment direction for the 7th layer of pythia-1.4b on the opening of Harry Potter in French.}
    
    \label{fig:neuroscope-many-texts}
\end{figure}

\section{Methods}
\subsection{Datasets and Models}\label{section:datasets}

\paragraph{ToyMovieReview}\label{dataset:toy-movie-review} A templatic dataset of continuation prompts we generated with the form 
\[\text{\tokens I thought this movie was ADJECTIVE, I VERBed it. Conclusion: This movie is}\]
where {\tokens ADJECTIVE } and {\tokens VERB } are either two positive words (e.g., {\tokens incredible} and {\tokens enjoyed}) or two negative words (e.g., {\tokens horrible} and {\tokens hated}) that are sampled from a fixed pool of 85 adjectives (split 55/30 for train/test) and 8 verbs. The expected completion for a positive review is one of a set of positive descriptors we selected from among the most common completions (e.g. {\tokens great}) and the expected completion for a negative review is a similar set of negative descriptors (e.g., {\tokens terrible}).

\paragraph{ToyMoodStory}\label{dataset:toy-mood-story} A similar toy dataset with prompts of the form \\[1ex]
{\tokens NAME1 VERB1 parties, and VERB2 them whenever possible. NAME2 VERB3 parties, and VERB4 them whenever possible. One day, they were invited to a grand gala. QUERYNAME feels very}\\[1ex]
To evaluate the model's output, we measure the logit difference between the ``{\tokens excited}" and ``{\tokens nervous}" tokens.

\paragraph{Stanford Sentiment Treebank (SST)}\label{dataset:stanford-sentiment-treebank} SST \cite{treebank} consists of 10,662 one sentence movie reviews with human annotated sentiment labels for every phrase from every review.

\paragraph{OpenWebText}\label{dataset:open-web-text} OWT \citep{Gokaslan2019OpenWeb} is the pretraining dataset for GPT-2 which we use as a source of random text for correlational evaluations.

\paragraph{GPT-2 and Pythia} \citep{radford2019language, biderman2023pythia} These are families of decoder-only transformer models with sizes varying from 85M to 2.8b parameters. We use GPT2-small for movie review continuation, pythia-1.4b for classification and pythia-2.8b for multi-subject tasks.

\subsection{Finding Directions}
\label{section:direction-finding-methods}

\begin{figure}
    \centering
    \includegraphics[width=0.5\linewidth,clip,trim={0 20cm 0 0}]{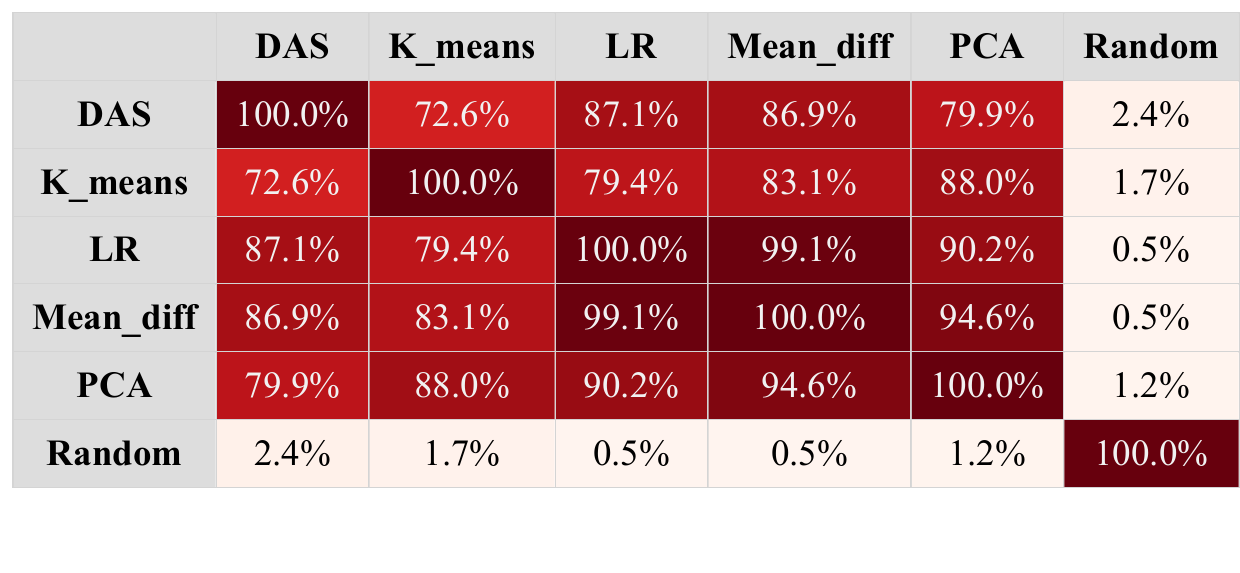}
    \caption{Cosine similarity of directions learned by different methods in GPT2-small's first layer. Each sentiment direction was derived from \textit{adjective} representations in the ToyMovieReview dataset (Section \ref{section:datasets}).}
    \label{fig:direction-similarities}
\end{figure}

We use five methods to find a sentiment direction in each layer of a language model using our ToyMovieReview dataset. In each of the following, let $\sP$ be the set of positive inputs and $\sN$ be the set of negative inputs. For some input $x \in \sP \cup \sN$, let $\va^L_{x}$ and $\vv_{x}^{L}$ be the vector in the residual stream at layer $L$ above the adjective and verb respectively. We reserve $\{\vv^L_{x}\}$ as a hold-out set for testing. Let the correct next token for $\sP$ be $p$ and for $\sN$ be $n$.

\paragraph{Mean Difference (MD)} The direction is computed as $\frac{1}{|\sP|} \sum_{p \in \sP} \va^L_{p} - \frac{1}{|\sN|} \sum_{n \in \sN} \va^L_{n}$.

\paragraph{$K$-means (KM)} We fit 2-means to $\{\va^L_{x}: x\in \sP\cup\sN\}$, obtaining cluster centroids $\{\vc_i: i \in [0, 1]\}$ and take the direction $\vc_1 - \vc_0$.

\paragraph{Linear Probing} The direction is the normed weights $\frac{\vw}{||\vw||}$ of a logistic regression (\textbf{LR}) classifier $\textbf{LR}(a_x^L) = \frac{1}{1+\exp(-\vw\cdot \va_x^L)}$ trained to distinguish between $x\in\sP$ and $x\in\sN$.

\paragraph{Distributed Alignment Search (DAS)}\label{method:das} \citep{das} The direction is a learned parameter $\theta$ where the training objective is the average logit difference 
\[ \sum_{x\in\sP} \left[ \textrm{logit}_\theta(x; p) - \textrm{logit}_\theta(x; n) \right] + \sum_{x\in\sN} \left[ \textrm{logit}_\theta(x; n) - \textrm{logit}_\theta(x; p) \right] \]
after patching using direction $\theta$ (see Section \ref{method:act-patching}).

\paragraph{Principal Component Analysis (PCA)} The direction is the first component of $\{\va^L_{x}: x \in \sP \cup \sN\}$.

\subsection{Causal Interventions}\label{section:causal-methods}

\paragraph{Activation Patching}\label{method:act-patching} In activation patching \citep{geiger-etal-2020-neural, vig2020causal}, we create two symmetrical datasets, where each prompt $x_\text{orig}$ and its counterpart prompt $x_\text{flipped}$ are of the same length and format but where key words are changed in order to flip the sentiment; e.g., ``This movie was great" could be paired with ``This movie was terrible." We first conduct a forward pass using $x_\text{orig}$ and capture these activations for the entire model. We then conduct forward passes using $x_\text{flipped}$, iteratively patching in activations from the original forward pass for each model component. We can thus determine the relative importance of various parts of the model with respect to the task currently being performed.
\cite{das} introduce distributed interchange interventions, a variant of activation patching that we call ``\gls{directional activation patching}". The idea is that rather than modifying the standard basis directions of a component, we instead only modify the component along a single direction in the vector space, replacing it during a forward pass with the value from a different input. 

We use two evaluation metrics. The logit difference (difference in logits for correct and incorrect answers) metric introduced in \cite{ioi}, as well as a ``logit flip" accuracy metric \citep{pmlr-v162-geiger22a}, which quantifies the proportion of cases where we induce an inversion in the predicted sentiment.


\paragraph{Ablations}\label{method:ablation} We eliminate the contribution of a particular component to a model's output, usually by replacing the component's output with zeros (zero-ablation) or the mean over some dataset (mean-ablation), in order to demonstrate its magnitude of importance. We also perform directional ablation, in which a component's activations are ablated only along a specific (e.g. sentiment) direction.

\section{Finding and Evaluating a `Sentiment Direction'} \label{section:direction-finding}

The first question we investigate is whether there exists a direction in the residual stream in a transformer model that represents the sentiment of the input text, as a special case of the linear representation hypothesis \citep{mikolov-etal-2013-linguistic}. We show that the methods discussed above (\ref{section:direction-finding-methods}) all arrive at a similar sentiment direction. Given some input text to a model, we can project the residual stream at a given token/layer onto a sentiment direction to get a `sentiment activation'.

\subsection{Visualizing and Comparing the Directions}
We fit directions using the ToyMovieReview dataset (Section \ref{section:datasets}) across various methods and finding extremely high cosine similarity between the learned sentiment directions (Figure \ref{fig:direction-similarities}).
This suggests that these are all noisy approximations of the same singular direction. Indeed, we generally found that the following results were very similar regardless of exactly how we specified the sentiment direction. The directions we found were not sparse vectors, as expected since the residual stream is not a privileged basis \citep{mathematicalframework}.

Here we show a visualisation in the style of Neuroscope \citep{nanda_neuroscope_io} where the projection is represented by color, with red being negative and blue being positive.
It is important to note that the direction being examined here was trained on just \simpleTrainSize positive and \simpleTrainSize negative English adjectives in an unsupervised way (using $K$-means with $K=2$). Notwithstanding, the extreme values along this direction appear readily interpretable in the wild in diverse text domains such as the opening paragraphs of Harry Potter in French (Figure \ref{fig:neuroscope-many-texts}). An interactive visualisation of residual stream directions in GPT2-small is available \href{http://ec2-34-192-101-140.compute-1.amazonaws.com:5014/}{here} \citep{yedidia2023website} and sentiment directions \href{https://github.com/curt-tigges/eliciting-latent-sentiment/tree/main/data/gpt2-small}{here}.

It is important to note that this type of analysis is qualitative, which should not act as a substitute for rigorous statistical tests as it is susceptible to interpretability illusions \citep{bolukbasi2021interpretability}. We rigorously evaluate our directions using correlational and causal methods.

\subsection{Correlational Evaluation}

\begin{figure}
    \centering
    \hfill
    \includegraphics[width=0.6\linewidth, clip, trim={0cm 0.37cm 0cm 2.7cm}]{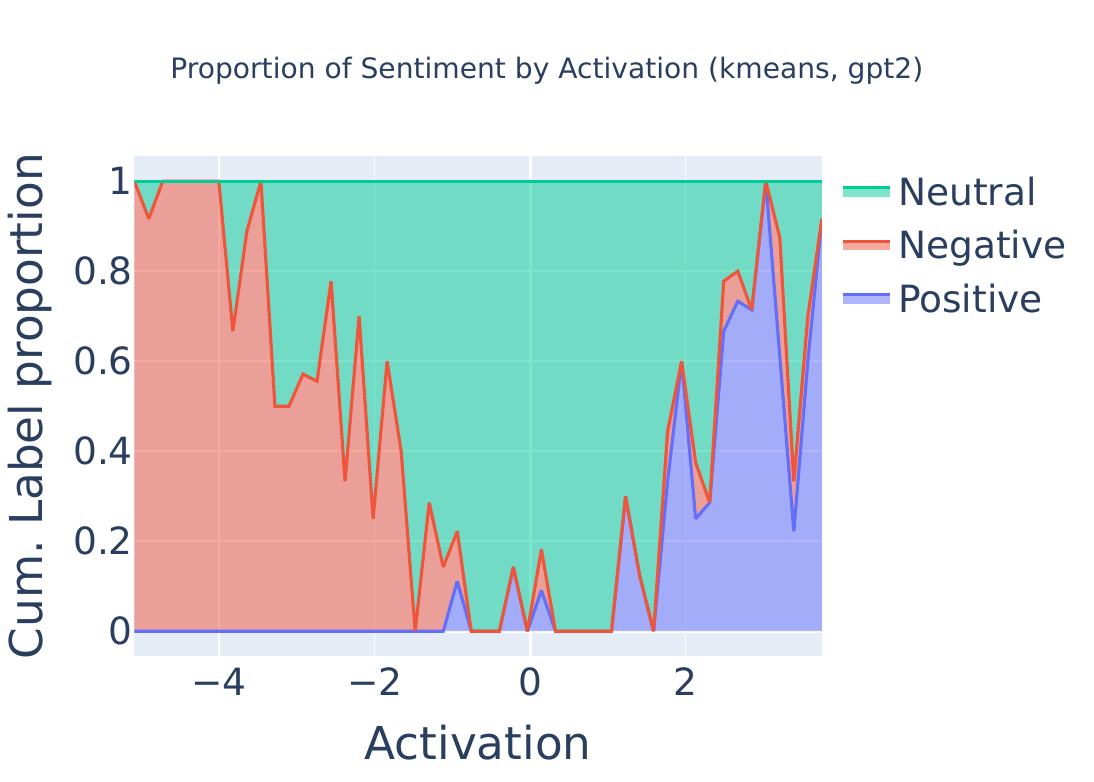}
    \hfill
    \resizebox{4cm}{!}{
    \small
    \renewcommand{\arraystretch}{1.7}
    \begin{tabular}[b]{cc}
\toprule
        \textbf{direction} & \textbf{accuracy}\\
        \midrule
        $K$-means & $78\%$\\
        PCA & $81\%$ \\
        Mean Diff & $80\%$\\
        LR & $89\%$\\
        DAS & $86\%$\\
        \bottomrule
    \end{tabular}
    }
    \hfill
    \caption{Area plot of sentiment labels for OpenWebText samples by $K$-means sentiment activation (left). Accuracy using sentiment activations to classify tokens as positive or negative (right). The threshold taken is the top/bottom 0.1\% of activations over OpenWebText. Sentiment activations are taken from GPT2-small's first residual stream layer. Classification was performed by GPT-4.}
    \label{tab:classification-accuracy}
\end{figure}

In a correlational analysis, we classify word sentiment by `sentiment activation' and show that the sentiment direction is sensitive to negation flipping sentiment.

\paragraph{Sentiment Directions Capture Lexical Sentiment} To test the meaning of the sentiment axis, we binned the sentiment activations of OpenWebText tokens from the first residual stream layer of GPT2-small into \owtNbins equal-width buckets and sampled \owtNsamples tokens from each. Then we asked GPT-4 to classify into Positive/Neutral/Negative. Specifically, we gave the GPT-4 API prompts of the following form: ``Your job is to classify the sentiment of a given token (i.e. word or word fragment) into Positive/Neutral/Negative. Token: `\{token\}'. Context: `\{context\}'. Sentiment:  " where the context length was \owtNcontext tokens centered around the sampled token. Only a cursory human sanity check was performed.

In Figure \ref{tab:classification-accuracy}, we show an area plot of the classifications by activation bin. We contrast the results for different methods in Table \ref{tab:classification-accuracy}. In the area plot we can see that the left side area is dominated by the ``Negative" label, whereas the right side area is dominated by the ``Positive" label and the central area is dominated by the ``Neutral" label. Hence the tails of the activations seem highly interpretable as representing a bipolar sentiment feature. The large space in the middle of the distribution simply occupied by neutral words (rather than a more continuous degradation of positive/negative) indicates superposition of features \citep{elhage2022superposition}. 

\paragraph{Negation Flips the Sentiment Direction} Using the $K$-means sentiment direction after the first layer of GPT2-small, we can obtain a view of how the model updates its view of sentiment during the forward pass, analogous to the ``logit lens`` technique from \cite{logit_lens}. In Figure \ref{fig:negation}, we see how the sentiment activation flips when the context of the sentiment word denotes that it is negated. Words like `fail', `doubt' and `uncertain' can be seen to flip from negative in the first couple of layers to being positive after a few layers of processing. An interesting task for future circuits analysis research could be to better understand the circuitry used to flip the sentiment axis in the presence of a negation context. We suspect significant MLP involvement (see Section \ref{section:interpreting-neurons}).

\subsection{Causal Evaluation}\label{section:direction-causal}

\begin{figure}
    \centering
    \includegraphics[width=0.95\linewidth,clip,trim={0 10cm 0 0}]{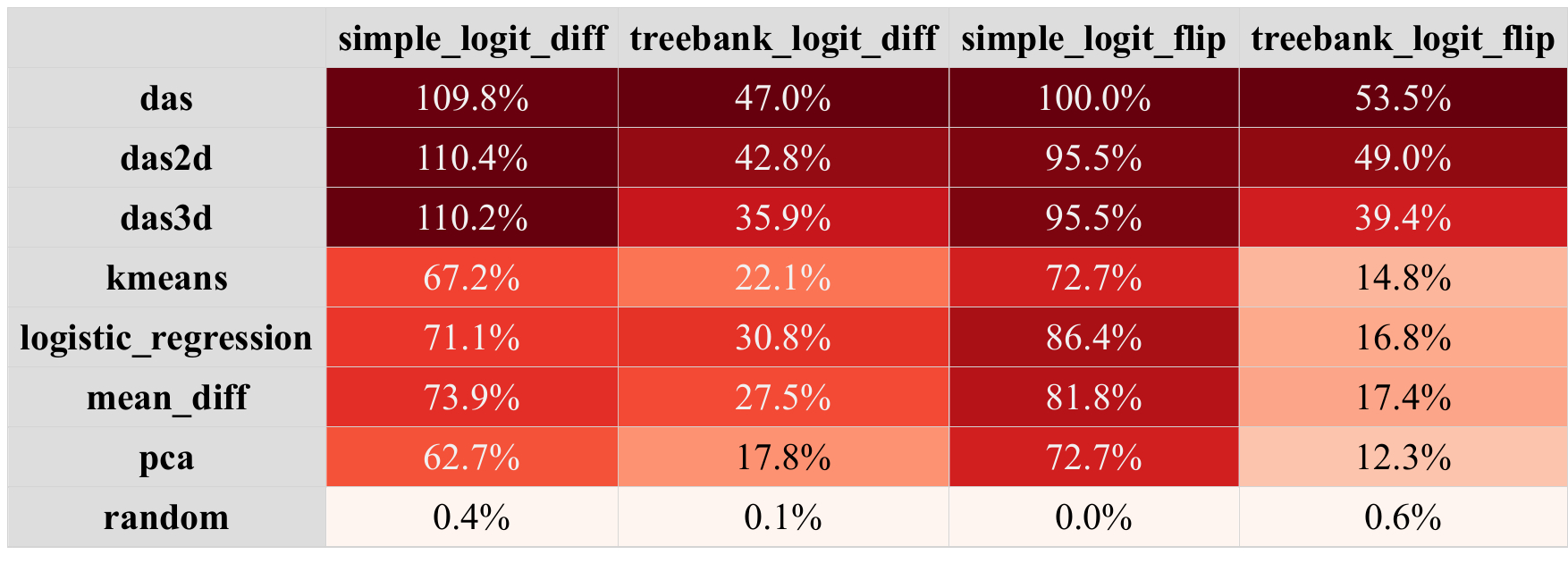}
    \caption{\Gls{directional patching} results for different methods in pythia-1.4b. We report the best result found across layers. The columns show two evaluation datasets, ToyMovieReview and Treebank, and two evaluation metrics, mean logit difference and \% of logit differences flipped.}
    \label{fig:direction-patching-simple}
\end{figure}

\paragraph{Sentiment directions are causal representations.} We evaluate the sentiment direction using \gls{directional patching} in Figure \ref{fig:direction-patching-simple}. These evaluations are performed on prompts with out-of-sample adjectives and the direction was not trained on \textit{any} verbs. Unsupervised methods such as $K$-means are still able to shift the logit differences and DAS is able to completely flip the prediction.

\begin{figure}
    \centering
    \includegraphics[width=\linewidth, clip, trim={0 0 0cm 0.95cm}]{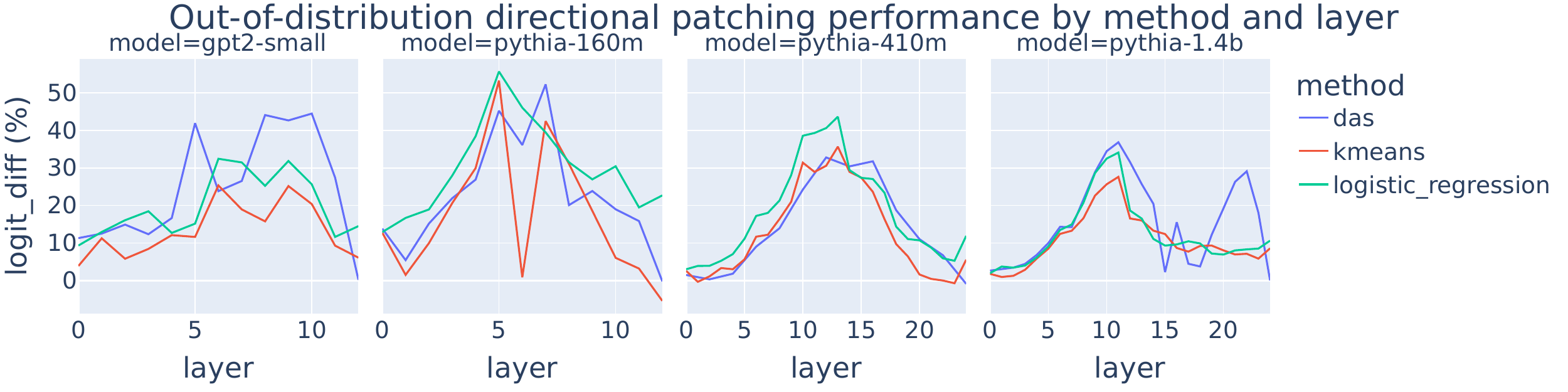}
    \caption{Patching results for directions trained on toy datasets and evaluated on the Stanford Sentiment Treebank test partition. We tend to find the best generalisation when training and evaluating at a layer near the middle of the model. We scaffold the prompt using the suffix {\tokens Overall the movie was very} and compute the logit difference between {\tokens good} and {\tokens bad}. The patching metric (y-axis) is then the \% mean change in logit difference.}
    \label{fig:direction-patching-layers}
\end{figure}

\paragraph{Directions Generalize Most at Intermediate Layers } If the sentiment direction was simply a trivial feature of the token embedding, then one might expect that \gls{directional patching} would be most effective in the first or final layer. However, we see in Figure \ref{fig:direction-patching-layers} that in fact it is in intermediate layers of the model where we see the strongest out-of-distribution performance to SST. This suggests the speculative hypothesis that the model uses the residual stream to form abstract concepts in intermediate layers and this is where the latent knowledge of sentiment is most prominent.

\paragraph{Activation Addition Steers the Model} A further verification of causality is shown in Figure \ref{fig:steering-proportions}. Here we use the technique of ``\gls{activation addition}" from \citet{turner2023activation}. We add a multiple of the sentiment direction to the first layer residual stream during each forward pass while generating sentence completions. Here we start from the baseline of a positive movie review: {\tokens ``I really enjoyed the movie, in fact I loved it. I thought the movie was just very..."}. By adding increasingly negative multiples of the sentiment direction, we find that indeed the completions become increasingly negative, without completely destroying the coherence of the model's generated text. We are wary of taking the model's activations out of distribution using this technique, but we believe that the smoothness of the transition in combination with the knowledge of our findings in the patching setting give us some confidence that these results are meaningful.
\paragraph{Validation on SST} We validate our sentiment directions derived from toy datasets (Section \ref{section:direction-causal}) on \gls{sst}. We collapsed the labels down to a binary ``Positive"/``Negative", just used the unique phrases rather than any information about their source sentences, restricted to the `test' partition and took a subset where pythia-1.4b can achieve 100\% zero shot classification accuracy, removing 17\% of examples. Then we paired up phrases of an equal number of tokens\footnote{We did this to maximise the chances of sentiment tokens occurring at similar positions} to make up 460 clean/corrupted pairs.
We used the scaffolding {\tokens ``Review Text: TEXT, Review Sentiment:"} and evaluated the logit difference between {\tokens ``Positive"} and {\tokens ``Negative"} as our \gls{patching metric}. Using the same DAS direction from Section \ref{section:direction-finding} trained on just a few examples and flipping the corresponding sentiment activation between clean/corrupted in a single layer, we can flip the output 53.5\% of the time (Figure \ref{fig:direction-patching-simple}).

\begin{figure}
    \centering
    \hfill
    \includegraphics[width=6cm, clip, trim={0.5cm 7cm 5.2cm 0}]{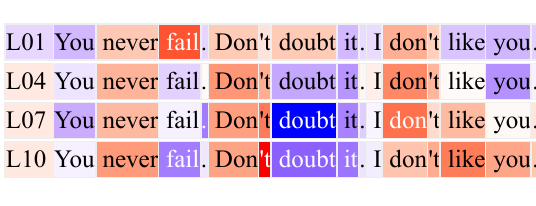}
    \hfill
    \small
    \begin{tabular}[b]{ccc}
        \toprule
        \textbf{direction} & \textbf{flip percent} & \textbf{flip median size}\\
        \midrule
        DAS & 96\% & 107\%\\
        KM & 96\% & 69\% \\
        MD & 89\% & 45\%\\
        LR & 100\% & 86\% \\
        PCA & 78\% & 44\% \\
        \bottomrule
    \end{tabular}
    \hfill
    \caption{We made a dataset of 27 negation examples and compute the change in sentiment activation at the negated token (e.g. {\tokens doubt}) between the 1st and 10th layers of GPT2-small.  We show sample text across layers for $K$-means (left), the fraction of activations flipped and the median size of the flip centered around the mean activation (right).}
    \label{tab:negation-experiment}
\end{figure}

\section{The Summarization Motif for Sentiment}\label{section:circuits}

\begin{figure}[t]
            \centering
            \includegraphics[width=0.7\linewidth,clip,trim={0cm 6cm 0cm 0cm}]{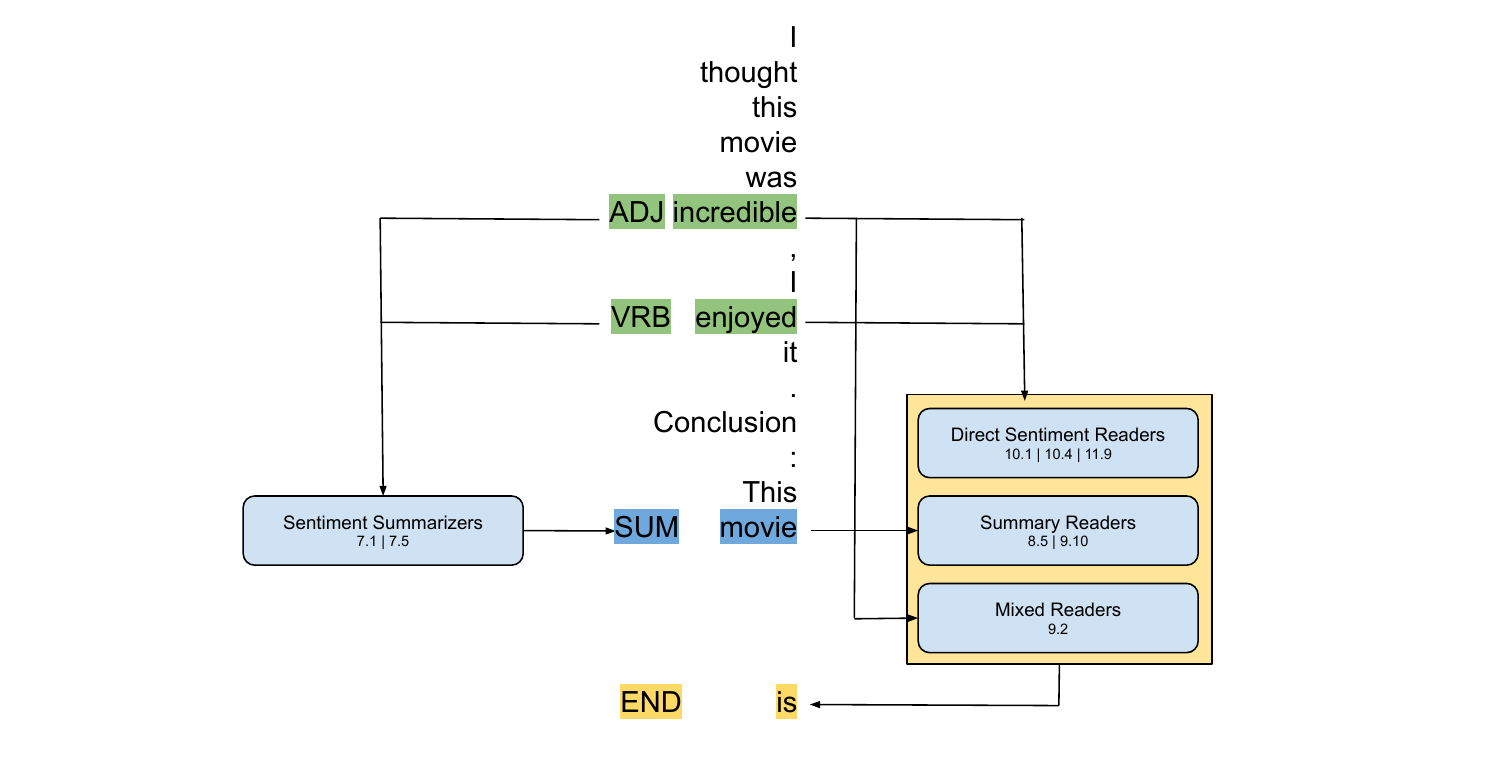}
            \caption{Primary components of GPT-2 sentiment circuit for the ToyMovieReview dataset. Here we can see both direct use of sentiment-laden words in predicting sentiment at {\tokens END} as well as an example of the summarization motif at the {\tokens SUM} position. Heads 7.1 and 7.5 write to this position and this information is causally relevant to the contribution of the summary readers at {\tokens END}.}
            \label{fig:movie-review-circuit-diagram}
        \end{figure}

    \subsection{Circuit Analyses}
    
    In this sub-section, we present circuit\footnote{We use the term ``circuit" as defined by \cite{ioi}, in the sense of a computational subgraph that is responsible for a significant proportion of the behavior of a neural network on some predefined task.}  analyses that give qualitative hints of the summarization motif, and restrict quantitative analysis of the summarization motif to \ref{section:summarization}.
Through an iterative process of path patching (see Section \ref{method:act-patching}) and analysing attention patterns, we have identified the circuit responsible for the ToyMovieReview task in GPT2-small (Figure \ref{fig:movie-review-circuit-diagram}) as well as the circuit for the ToyMoodStories task.
Below, we provide a brief overview of the circuits we identified, reserving the full details for \ref{section:detailed-circuit-analysis}.

        \paragraph{Initial observations of summarization in GPT-2 circuit for ToyMovieReview} Mechanistically, this is a binary classification task, and a naive hypothesis is that attention heads attend directly from the final token to the valenced tokens and map positive sentiment to positive outputs and vice versa. This happens, but in addition attention head output is causally important at intermediate token positions, which are then read from when producing output at {\tokens END}. We consider this an instance of summarization, in which the model aggregates causally-important information relating to an entity at a particular token for later usage, rather than simply attending back to the original tokens that were the source of the information. 
        
        We find that the model performs a simple, interpretable algorithm to perform the task (using a circuit made up of 9 attention heads):

        \begin{enumerate}[leftmargin=0.5cm]
            \item Identify sentiment-laden words in the prompt, at {\tokens ADJ} and {\tokens VRB}.
            \item Write out sentiment information to SUM (the final ``movie" token).
            \item Read from {\tokens ADJ}, {\tokens VRB} and {\tokens SUM} and write to {\tokens END}.\footnote{We note that our patching experiments indicate that there is no causal dependence on the output of other model components at the ADJ and VRB positions--only at the SUM position.}
        \end{enumerate}

        The results of activation patching the residual stream can be seen in the Appendix, Fig. \ref{fig:toy-movie-review-circuit-patching}. The output of attention heads is only important at the {\tokens movie} position, which we designate as {\tokens SUM}. We label these heads ``sentiment summarizers." Specific attention heads attend to and rely on information written to this token position as well as to {\tokens ADJ} and {\tokens VRB}.
        
        To validate this circuit and the involvement of the sentiment direction, we patched the entirety of the circuit at the {\tokens ADJ} and {\tokens VRB} positions along the sentiment direction only, achieving a 58.3\% rate of logit flips and a logit difference drop of 54.8\% (in terms of whether a positive or negative word was predicted). Patching the circuit at those positions along all directions resulted in flipping 97\% of logits and a logit difference drop of 75\%, showing that the sentiment direction is responsible for the majority of the function of the circuit.
        \paragraph{Multi-subject mood stories in Pythia 2.8b}\label{section:pythia-2.8b-circuit}
        \begin{figure}[h]
            \centering
            \includegraphics[width=0.8\linewidth]{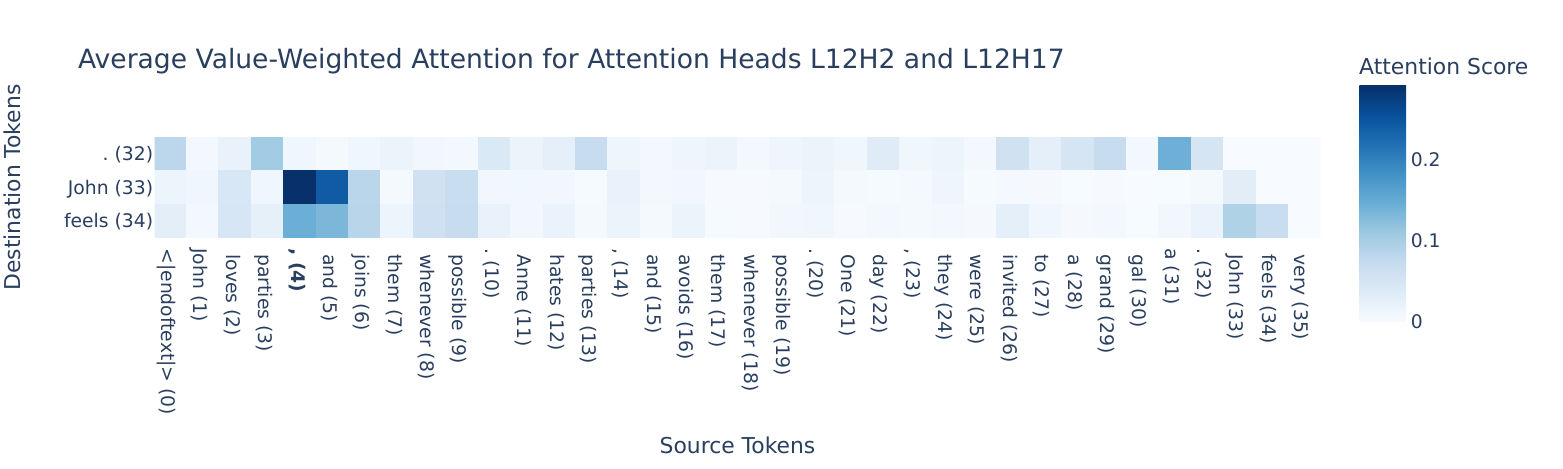} %
            \caption[caption for LOF]{Value-weighted\footnotemark averaged attention to commas and comma phrases in Pythia-2.8b from the top two attention heads writing to the repeated name and ``feels" tokens--two key components of the summarization sub-circuit in the ToyMoodStories task. Note that they attend heavily to the relevant comma from both destination positions.}
            \label{fig:pythia-2.8b-attn_to_commas}
        \end{figure}
        
        \footnotetext{That is, the attention pattern weighted by the norm of the value vector at each position as per \cite{kobayashi2020attention}. We favor this over the raw attention pattern as it filters for \textit{significant} information being moved.}
        
        We next examined the circuit that processes the mood dataset in Pythia-2.8b (the smallest model that could perform the task), which is a more complex task that requires more summarization. As such it presents a better object for study of this motif. We reserve a detailed description of the circuit for the Appendix, but here we observed increasing reliance on summarization, specifically:
        \begin{itemize}[leftmargin=0.5cm]
            \item A set of attention heads \textbf{attended primarily to the comma} following the preference phrase for the queried subject (e.g. {\tokens John hates parties,}), and secondarily to other words in the phrase, as seen in Figure \ref{fig:pythia-2.8b-attn_to_commas}. We observed this phenomenon both with regular attention and value-weighted attention, and found via path patching that \textbf{these heads relied partially on the comma token} for their function, as seen in Figure \ref{fig:pythia-2.8b-path_patching_commas}.
            \item Heads attending to preference phrases (both commas and other tokens) tended to write to the repeated name token near the end of the sentence ({\tokens John}) as well as to the {\tokens feels} token--another type of summarization behavior. Later heads attended to the repeated name and {\tokens feels} tokens with an output important to {\tokens END}.
        \end{itemize}
    
    \subsection{Exploring and validating summarization behavior in punctuation}\label{section:summarization}
    Our circuit analyses reveal suggestive evidence that summarization behavior at intermediate tokens like commas, periods and certain nouns plays an important part in sentiment processing, despite these tokens having no inherent valence. We focus on summarization at commas and periods and explore this further in a series of ablation and patching experiments. We find that in many cases this summarization results in a partial information bottleneck, in which the summarization points become as important (or sometimes more important) than the phrases that precede them for sentiment tasks.
        \paragraph{Summarization information is comparably important as original semantic information}
        In order to determine the extent of the information bottleneck presented by commas in sentiment processing, we tested the model's performance on the multi-subject mood stories dataset mentioned above. We froze the model's attention patterns to ensure the model used the information from the patched commas in exactly the same way as it would have used the original information. Without this step, the model could simply avoid attending to the commas. We then performed activation patching on either the precomma phrases (e.g., patching ``{\tokens John hates parties,}" with ``{\tokens John loves parties,}") while freezing the commas so they retain their original, unflipped values; or on the two commas alone, and find a similar drop in the logit difference for both as shown in table \ref{table:patching-commas-vs-precommas}.

        \begin{table}[h]
            \caption{Patching results at summary positions}
            \label{tab:comma-and-period-patching}
            \centering
            \begin{subtable}{0.45\textwidth}
            \resizebox{0.8\textwidth}{!}{
            \begin{tabular}{|c|c|}
            \hline
            \textbf{Intervention} & \textbf{Change in } \\ 
             & \textbf{logit difference} \\ \hline
            Patching full phrase   & -75\% \\ 
            values (incl. commas) & \\ \hline
            Patching pre-comma  & -38\% \\
             values (freezing commas) &  \\ \hline
            Patching comma values only & -37\% \\ \hline
            \end{tabular}
            }
            \caption{Change in logit difference from intervention on attention head value vectors}
            \label{table:patching-commas-vs-precommas}
            \end{subtable}
            \hfill
            \begin{subtable}{0.48\textwidth}
            \resizebox{0.8\textwidth}{!}{
            \begin{tabular}{|c|c|}
            \hline
            \textbf{Count of irrelevant tokens } & \textbf{Ratio of LD change} \\ 
            \textbf{after preference phrase} & \textbf{for periods vs. phrases} \\ \hline
            0 tokens & 0.29 \\ \hline
            10 tokens & 0.63 \\ \hline
            18 tokens & 0.92  \\ \hline
            22 tokens & 1.15  \\ \hline
            \end{tabular}
            }
            \caption{Ratio between logit difference change for periods vs. pre-period phrases after patching values}
            \label{table:ld-ratio-by-length}
            \end{subtable}
        \end{table}
        
        \paragraph{Importance of summarization increases with distance}
        We also observed that reliance on summarization tends to increase with greater distances between the preference phrases and the final part of the prompt that would reference them. To test this, we injected irrelevant text\footnote{E.g. ``John loves parties. \textit{He has a red hat and wears it everywhere, especially when he is riding his bicycle through the city streets.} Mark hates parties. \textit{He has a purple hat but only wears it on Sundays, when he takes his weekly walk around the lake.} One day, they were invited to a grand gala. John feels very"} after each of the preference phrases in our multi-subject mood stories (after "John loves parties." etc.) and measured the ratio between logit difference change for the periods at the end of these phrases vs. pre-period phrases, with higher values indicating more reliance on period summaries (Table \ref{table:ld-ratio-by-length}). We found that the periods can be up to \maxLDratio \textbf{more} important than the actual phrases as this distance grows. Although these results are only a first step in assessing the importance of summarization importance relative to prompt length, our findings suggest that this motif may only increase in relative importance as models grow in context length, and thus merits further study.

    \subsection{Validating summarization behavior in SST}\label{section:summarization-treebank}
    In order to study more rigorously how summarization behaves with natural text, we examined this phenomenon in SST. We appended the suffix ``Review Sentiment:" to each of the prompts and evaluate Pythia-2.8b on zero-shot classification according to whether {\tokens positive} or {\tokens negative} have higher probability and are in the top 10 tokens predicted.  We then take the subset of examples Pythia-2.8b succeeds on that have at least one comma, which means we start with a baseline of 100\% accuracy. We performed ablation and patching experiments on comma representations. If comma representations do not summarize sentiment information, then our experiments should not damage the model's abilities. However, our results reveal a clear summarization motif for SST.
            \paragraph{Ablation baselines}
            We performed two baseline experiments in order to obtain a control for our later experiments. First to measure the total effect of the sentiment directions, we performed directional ablation (as described in \ref{method:act-patching}) using the sentiment directions found with DAS to every token at every layer, resulting in a \ablateDirectionTreebankLDDrop reduction in the logit difference and a \ablateDirectionTreebankAccDrop drop in accuracy (to \ablateDirectionTreebankAcc). Second, we performed directional ablation on all tokens with a small set of random directions, resulting in a $<1\%$ change to the same metrics.
            \paragraph{Directional ablation at all comma positions} We then performed directional ablation--using the DAS (\ref{method:das}) sentiment direction--to every comma in each prompt, regardless of position, resulting in an \ablateCommasDirectionTreebankLDDrop drop in the logit difference and an \ablateCommasDirectionTreebankAccDrop drop in zero-shot classification accuracy--indicating that nearly 50\% of the model's sentiment-direction-mediated ability to perform the task accurately was mediated via sentiment information at the commas. We find this particularly significant because we did not take any special effort to ensure that commas were placed at the end of sentiment phrases.
            \paragraph{Mean-ablation at all comma positions} Separately from the above, we performed mean ablation at all comma positions as in \ref{method:ablation}, replacing each comma activation vector with the mean comma activation from the entire dataset in a layerwise fashion. Note that this changes the entire activation on the comma token, not just the activation in the sentiment direction. This resulted in a \ablateCommasMeanTreebankLDDrop drop in logit difference and an accuracy drop of \ablateCommasMeanPatchTreebankAccDrop.

    \subsection{The big picture of summarization}
    We have identified a phenomenon across multiple models and tasks where sentiment information is not directly transferred from valenced tokens to the final output but is first aggregated at intermediate, non-valenced tokens like commas and periods (and sometimes noun tokens for specific referents). We call this behavior the ``summarization motif." These summarization points serve as partial information bottlenecks and are causally significant for the model's performance on sentiment tasks. Through a series of ablation and patching experiments, we have validated the importance of this summarization behavior in both toy tasks and real-world datasets like the Stanford Sentiment Treebank. Additional findings suggest that as models grow in context length, the importance of this internal summarization behavior may increase--a subject that warrants further investigation. Overall, the discovery of this summarization behavior adds a new layer of complexity to our understanding of how sentiment is processed and represented in LLMs, and seems likely to be an important part of how LLMs create internal world representations.

\section{Related Work}\label{related work}

\paragraph{Sentiment Analysis} Understanding the emotional valence in text data is one of the first NLP tasks to be revolutionized by deep learning \citep{treebank} and remains a popular task for benchmarking NLP models \citep{rosenthal-etal-2017-semeval, nakov-etal-2016-semeval, potts-etal-2020-dynasent, abraham-etal-2022-cebab}.  For a review of the literature, see \citep{Pang:2008,Liu_2012, Grimes:2014}. 

\paragraph{Understanding Internal Representations} 
This research was inspired by the field of Mechanistic Interpretability, an agenda which aims to reverse-engineer the learned algorithms inside models \citep{olah2020zoom, mathematicalframework, nanda2023grokking}. Exploring representations (Section \ref{section:direction-finding}) and world-modelling behavior inside transformers has garnered significant recent interest. This was studied in the context of synthetic game-playing models by \citet{li2023emergent} and evidence of linearity was demonstrated by \cite{nanda_othello_2023} in the same context. Other work studying examples of world-modelling inside neural networks includes \citet{li2021alchemy, patel2022mapping, abdou2021language}. Another framing of a very similar line of inquiry is the search for latent knowledge \citep{christiano2021eliciting, burns2022discovering}. Prior to the transformer, representations of emotion were studied in \citet{goh2021multimodal} and sentiment was studied by \citet{radford2017lsentiment}, notably, the latter finding a sentiment neuron which implies a linear representation of sentiment. A linear representation of truth in LLMs was found by \citet{marks2023geometry}.

\paragraph{Summarization Motif} Our study of the Summarization motif (Section \ref{section:circuits}) follows from the search for information bottlenecks in models (\cite{li2021alchemy}). Our use of the word `motif', in the style of \citet{olah2020zoom}, is originally inspired from systems biology \citep{alon2006motifs}. The idea of exploring representations at different frequencies or levels of abstraction was explored further in \citet{tamkin2020language}. Information storage after the relevant token was observed in how GPT2-small predicts gender \citep{gendered_pronouns}.

\paragraph{Causal Interventions in Language Models} 
We approach our experiments from a causal mediation analysis perspective. 
Our approach to identifying computational subgraphs that utilize feature representations as inspired by the `circuits analysis' framework \citep{heimersheim2023PythonDocstrings, deepmind2023circuit, hanna2023GreaterThan}, especially the tools of \gls{mean ablation} and \gls{activation patching}  \citep{vig2020causal, Geiger:Lu-etal:2021, Geiger-etal:2023:CA, rome, ioi, acdc, scrubbing, gevaFact}. We use Distributed Alignment Search \citep{das} in order to apply these ideas to specific subspaces.

\section{Conclusion}\label{section:discussion}
The two central novel findings of this research are the existence of a linear representation of sentiment and the use of summarization to store sentiment information. We have seen that the sentiment direction is causal and central to the circuitry of sentiment processing. Remarkably, this direction is so stark in the residual stream space that it can be found even with the most basic methods and on a tiny toy dataset, yet generalise to diverse natural language datasets from the real-world. Summarization is a motif present in larger models with longer context lengths and greater proficiency in zero-shot classification. These summaries present a tantalising glimpse into the world-modelling behavior of transformers. 

We also see this research as a model for how to find and study the representation of a particular feature. Whereas in dictionary learning \citep{bricken2023monosemanticity} we enumerate a large set of features which we then need to interpret, here we start with an interpretable feature and subsequently verify that a representation of this feature exists in the model, analogously to \citet{zou2023representation}. One advantage of this is that our fitting process is much more efficient: we can use toy datasets and very simple fitting methods. It is therefore very encouraging to see that the results of this process generalise well to the full data distribution, and indeed we focus on providing a variety of experiments to strengthen the case for the existence of our hypothesised direction.

\paragraph{Limitations}\label{section:limitations}
Did we find a truly universal sentiment direction, or merely the first principal component of directions used across different sentiment tasks? As found by \citet{bricken2023monosemanticity}, we suspect that this feature could be ``split" further into more specific sentiment features.

Similarly, one might wonder if there is really a single bipolar sentiment direction or if we have simply found the difference between a ``positive" and a ``negative" sentiment direction. It turns out that this distinction is not well-defined, given that we find empirically that there is a direction corresponding to ``valenced words". Indeed, if $\vx$ is the valence direction and $\vy$ is the sentiment direction, then $\vp=\vx+\vy$ represents positive sentiment and $\vn=\vx-\vy$ is the negative direction. Conversely, we can reframe as starting from the positive/negative directions $\vp$ and $\vn$, and then re-derive $\vx=\frac{\vp+\vn}{2}$ and $\vy:=\frac{\vp-\vn}{2}$.

Many of our casual abstractions do not explain 100\% of sentiment task performance. There is likely circuitry we've missed, possibly as a result of distributed representations or superposition \citep{elhage2022superposition} across components and layers. This may also be a result of self-repair behavior \citep{ioi, mcgrath2023hydra}. Patching experiments conducted on more diverse sentence structures could also help to better isolate the circuitry for sentiment from more task-specific machinery.

The use of small datasets versus many hyperparameters and metrics poses a constant risk of gaming our own measures. Our results on the larger and more diverse \gls{sst} dataset, and the consistent results across a range of models help us to be more confident in our results.

Distributed Alignment Search (\gls{das}) outperformed on most of our metrics but presents possible dangers of overfitting to a particular dataset and taking the activations out of distribution \citep{lange2023interpretability}. We include simpler tools such as Logistic Regression as a sanity check on our findings. Ideally, we would love to see a set of best practices to avoid such illusions.

\paragraph{Implications and future work}
The summarization motif emerged naturally during our investigation of sentiment, but we would be very interested to study it in a broader range of contexts and understand what other factors of a particular model or task may influence the use of summarization.

When studying the circuitry of sentiment, we focused almost exclusively on attention heads rather than MLPs. However, early results suggest that further investigation of the role of MLPs and individual neurons is likely to yield interesting results (\ref{section:interpreting-neurons}).

Finally, we see the long-term goal of this line of research as being able to help detect dangerous computation in language models such as \textit{deception}. Even if the existence of a single ``deception direction" in activation space seems a bit naive to postulate, hopefully in the future many of the tools developed here will help to detect representations of deception or of knowledge that the model is concealing, helping to prevent possible harms from LLMs.



\subsubsection*{Author Contributions}
Oskar and Curt made equal contributions to this paper. Curt's focus was on circuit analysis and he discovered the summarization motif, leading to Section \ref{section:circuits}. Oskar was focused on investigating the direction and eventually conducted enough independent experiments to convince us that the direction was causally meaningful, leading to Section \ref{section:direction-finding}. Neel was our mentor as part of SERI MATS, he suggested the initial project brief and provided considerable mentorship during the research. He also did the neuron analysis in Section \ref{section:interpreting-neurons}. Atticus acted a secondary source of mentorship and guidance. His advice was particularly useful as someone with more of a background in causal mediation analysis. He suggested the use of Stanford Sentiment Treebank and the discrete accuracy metric.

\subsubsection*{Acknowledgments}
SERI MATS provided funding, lodging and office space for 2 months in Berkeley, California. The transformer-lens package \citep{nanda2022transformerlens} was indispensable for this research. We are very grateful to Alex Tamkin for his extensive feedback. Other valuable feedback came from Georg Lange, Alex Makelov and Bilal Chughtai.  Atticus Geiger is supported by a grant from Open Philanthropy.

\section*{Reproducibility Statement}
To facilitate reproducibility of the results presented in this paper, we have provided detailed descriptions of the datasets, models, training procedures, algorithms, and analysis techniques used. The ToyMovieReview dataset is fully specified in Section \ref{section:toy-movie-review-details}. We use publicly available models including GPT-2 and Pythia, with details on the specific sizes provided in Section 2.1. The methods for finding sentiment directions are described in full in Section 2.2. Our causal analysis techniques of activation patching, ablation, and directional patching are presented in Section 2.3. Circuit analysis details are extensively covered for two examples in Appendix Section A.3. The code for data generation, model training, and analyses is available \href{https://github.com/curt-tigges/eliciting-latent-sentiment}{here}.

\bibliography{iclr2024}
\bibliographystyle{iclr2024}

\newpage\appendix
\renewcommand\thefigure{\thesection.\arabic{figure}}  
   
\section{Appendix}\label{section:appendix}
\setcounter{figure}{0} 

\begin{figure}[t]
    \centering
    
    \begin{subfigure}[b]{0.49\linewidth}
    \caption{PCA on adjectives in and out of sample}
    \label{fig:pca_oos_adj}
        \centering
        \includegraphics[width=\linewidth]{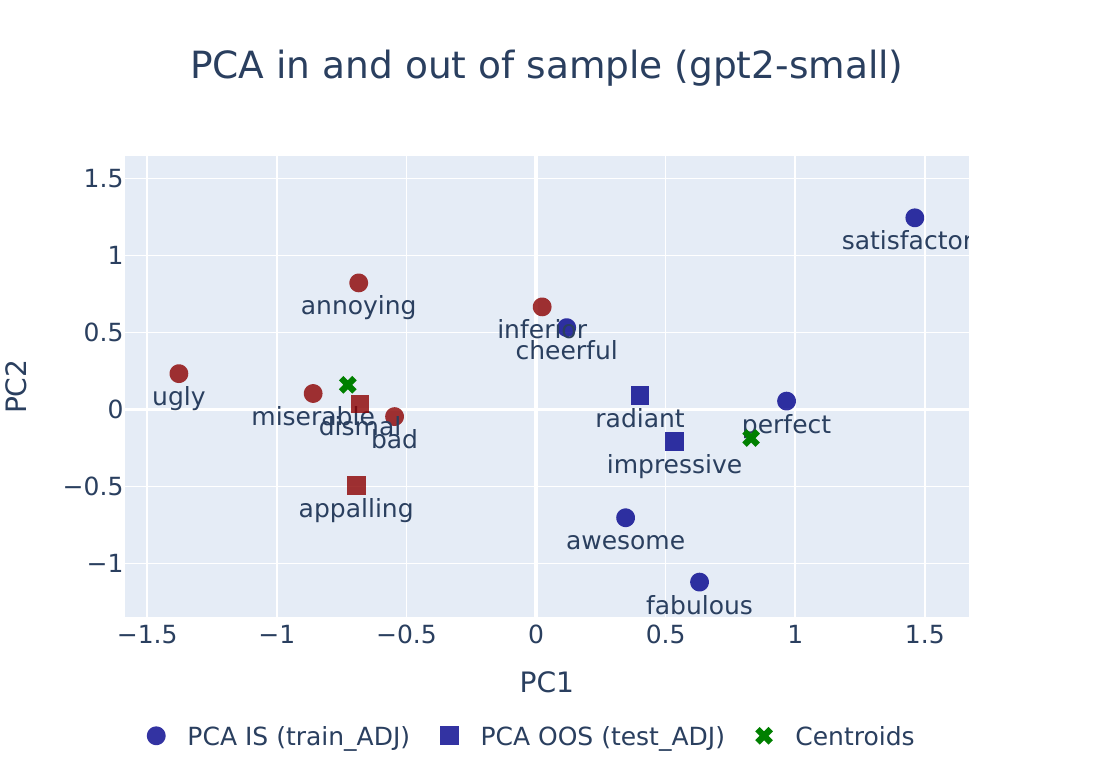}
    \end{subfigure}
    \hfill
    \begin{subfigure}[b]{0.49\linewidth}
    \caption{PCA on in-sample adjectives and out-of-sample verbs}
    \label{fig:pca_ood_verbs}
        \centering
        \includegraphics[width=\linewidth]{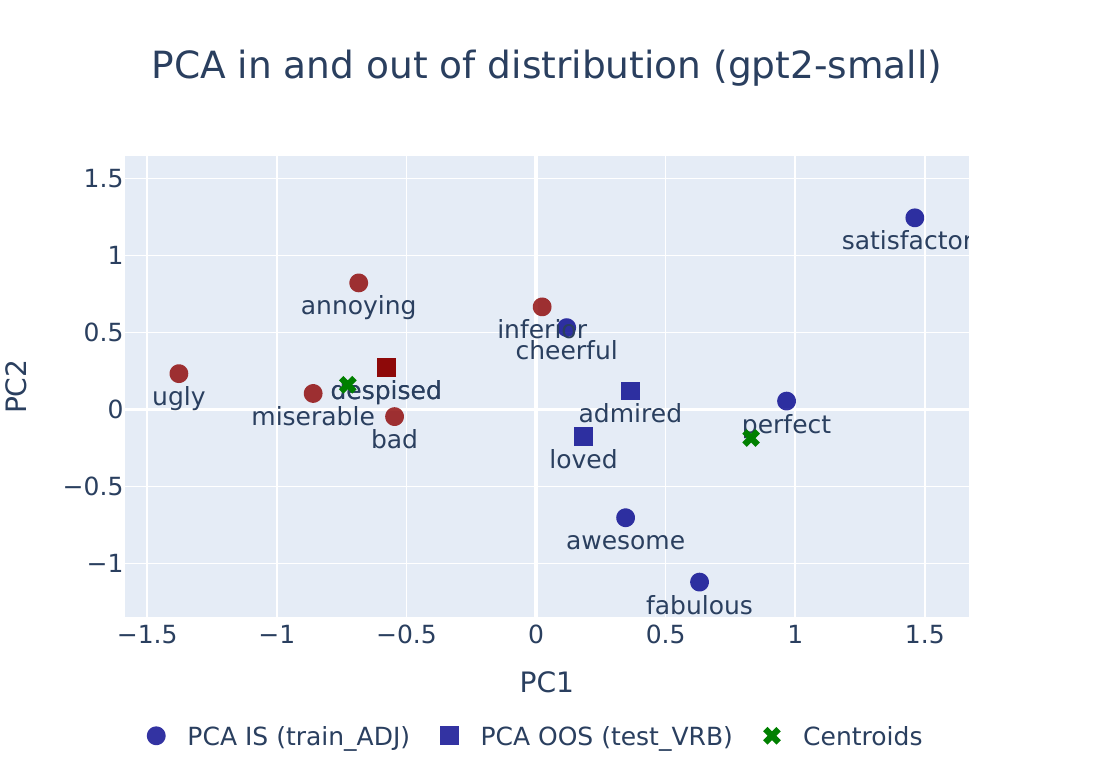}
    \end{subfigure}
    
    \caption{2-D PCA visualization of the embedding for a handful of adjectives and verbs (GPT2-small)}
    \label{fig:pca_2d}
\end{figure}

\subsection{Further evidence for a linear sentiment representation}

\subsubsection{Clustering}\label{section:pca-kmeans}

In Section \ref{section:direction-finding-methods}, we outline just a few of the many possible techniques for determining a direction which hopefully corresponds to sentiment. Is it overly optimistic to presume the existence of such a direction? The most basic requirement for such a direction to exist is that the residual stream space is clustered. We confirm this in two different ways.

First we fit 2-D PCA to the token embeddings for a set of \simpleTrainSize positive and \simpleTrainSize negative adjectives. In Figure \ref{fig:pca_2d}, we see that the positive adjectives (blue dots) are very well clustered compared to the negative adjectives (red dots). Moreover, we see that sentiment words which are out-of-sample with respect to the PCA (squares) also fit naturally into their appropriate color. This applies not just for unseen adjectives (Figure \ref{fig:pca_oos_adj}) but also for verbs, an entirely out-of-distribution class of word (Figure \ref{fig:pca_ood_verbs}). 

\begin{figure}[ht]
    \centering

    \begin{subfigure}{0.5\linewidth}
        \centering
        \includegraphics[width=0.9\linewidth,clip,trim={0 5cm 1cm 0}]{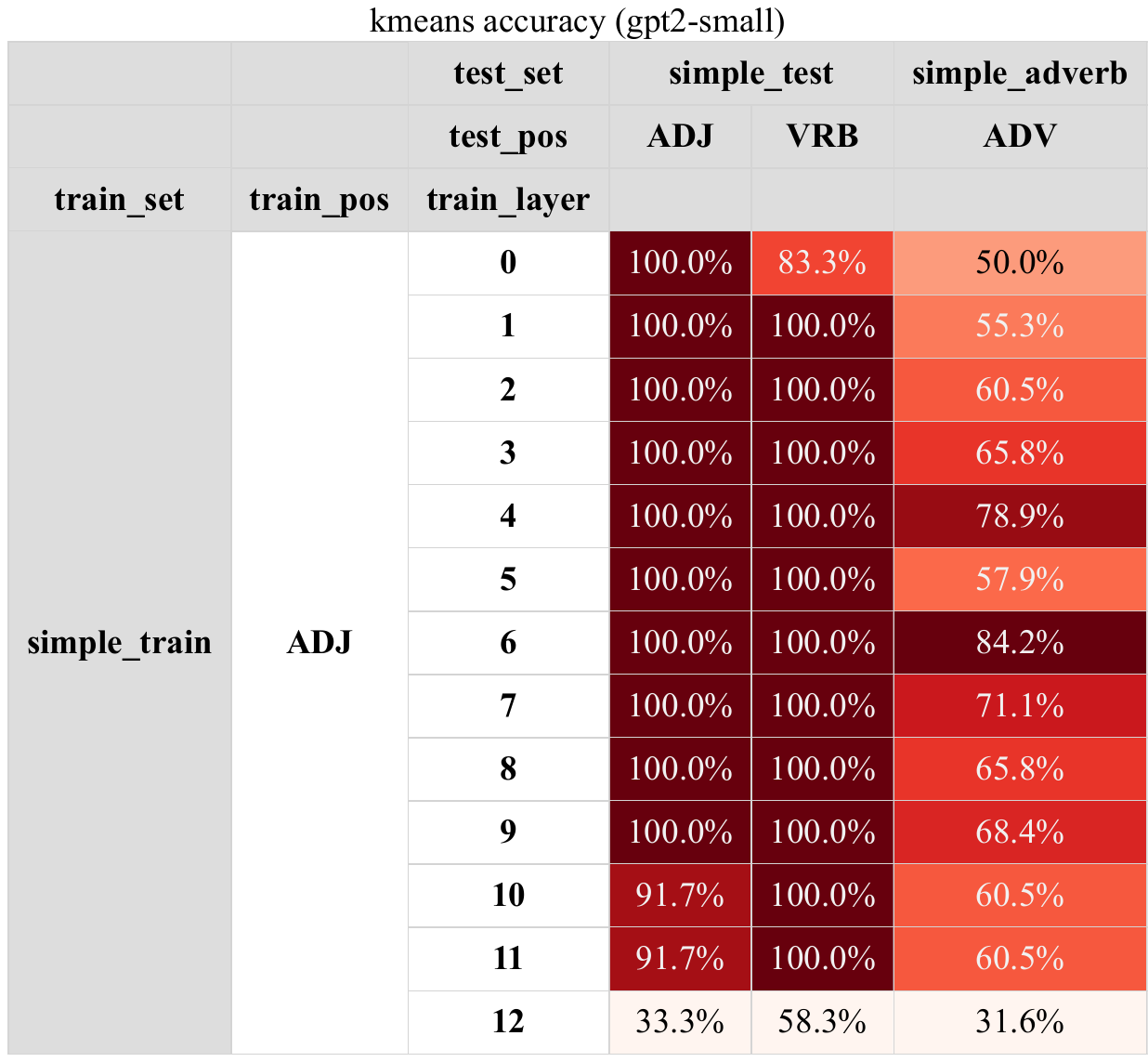}
        \caption{GPT-2 Small}
        \label{fig:kmeans-accuracy-gpt2-small}
    \end{subfigure}%
    \begin{subfigure}{0.5\linewidth}
        \centering
        \includegraphics[width=0.9\linewidth,clip,trim={0 8cm 8cm 0}]{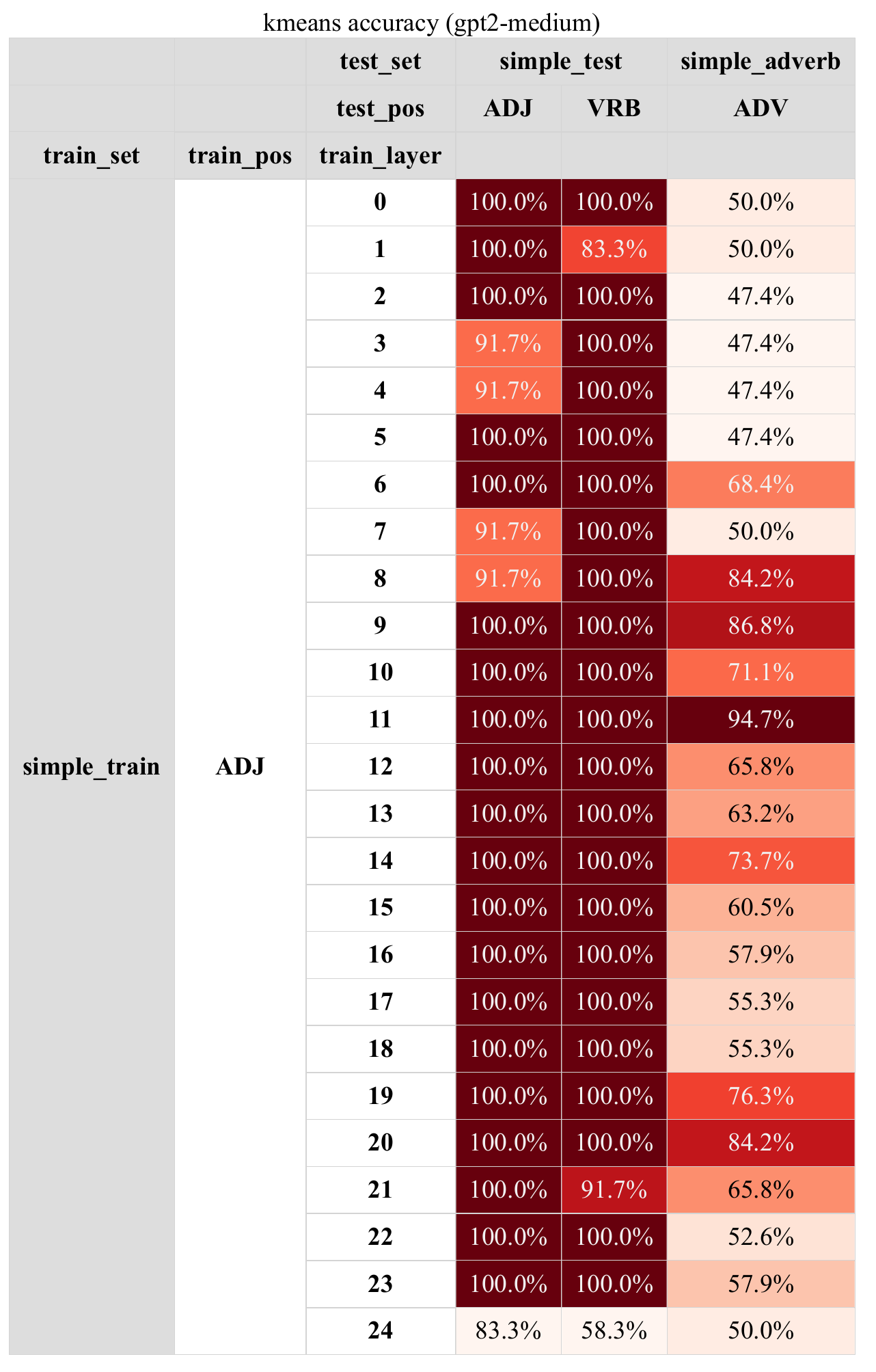}
        \caption{GPT-2 Medium}
        \label{fig:kmeans-accuracy-gpt2-medium}
    \end{subfigure}
    
    \begin{subfigure}{0.5\linewidth}
        \centering
        \includegraphics[width=0.9\linewidth,clip,trim={0 8.5cm 8cm 0}]{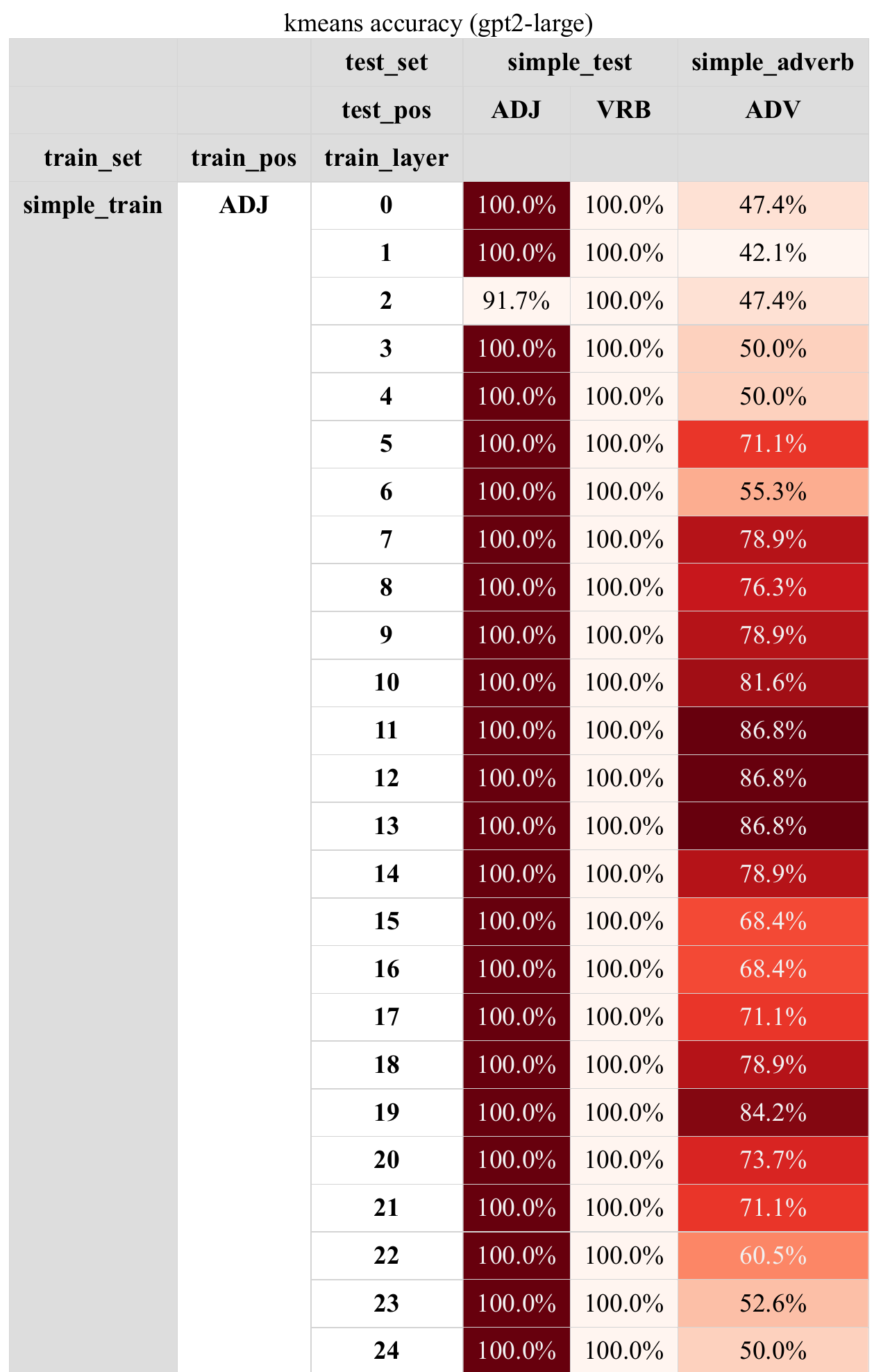}
        \caption{GPT-2 Large}
        \label{fig:kmeans-accuracy-gpt2-large}
    \end{subfigure}%
    \begin{subfigure}{0.5\linewidth}
        \centering
        \includegraphics[width=0.9\linewidth,clip,trim={0 8.5cm 8cm 0}]{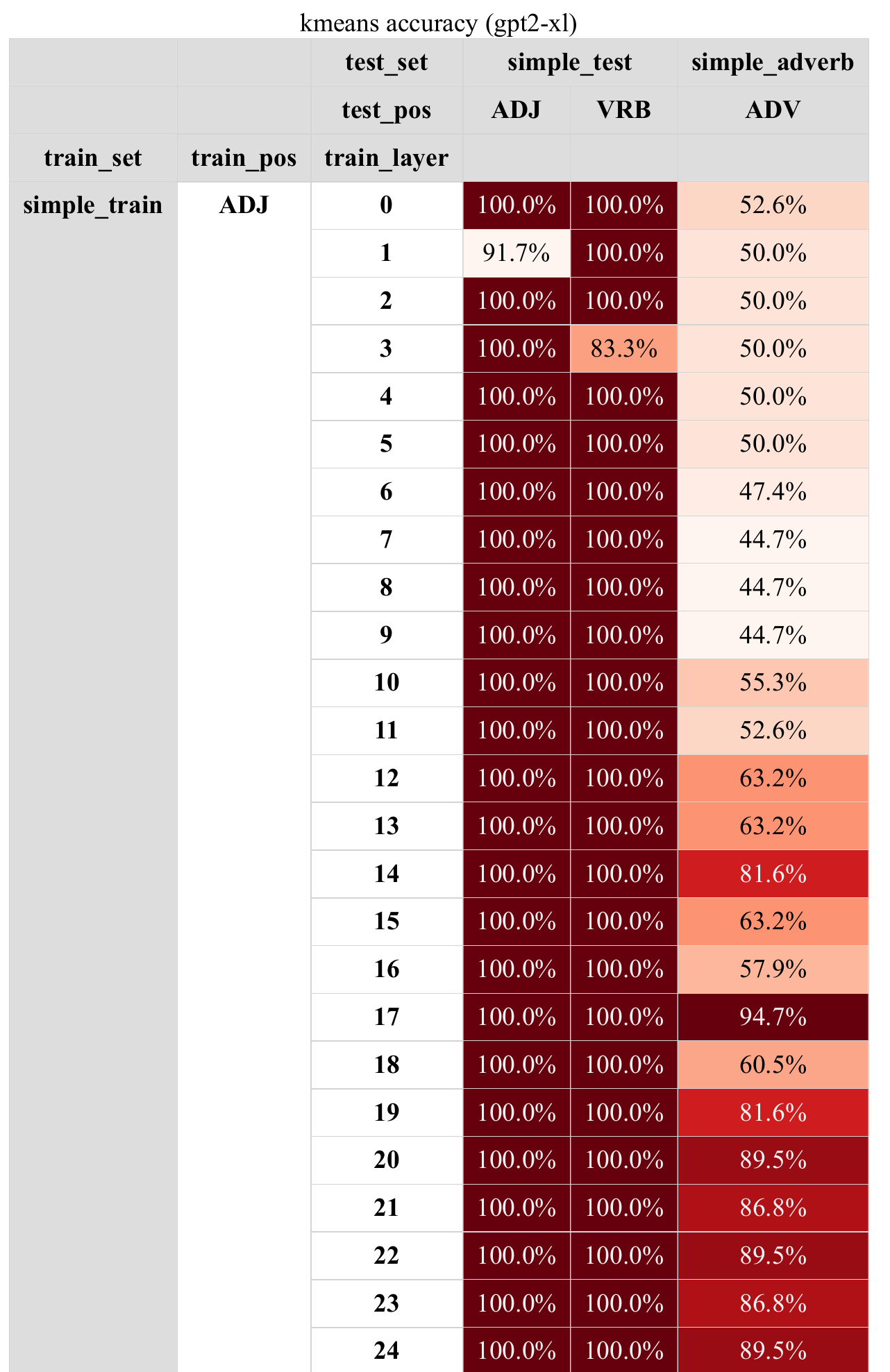}
        \caption{GPT-2 XL}
        \label{fig:kmeans-accuracy-gpt2-xl}
    \end{subfigure}
    
    \caption{2-means classification accuracy for various GPT-2 sizes, split by layer (showing up to 24 layers)}
    \label{fig:kmeans-accuracy-gpt2}
\end{figure}

Secondly, we evaluate the accuracy of 2-means trained on the Simple Movie Review Continuation adjectives (Section \ref{section:datasets}). The fact that we can classify in-sample is not very strong evidence, but we verify that we can also classify out-of-sample with respect to the $K$-means fitting process. Indeed, even on hold-out adjectives and on the verb tokens (which are totally out of distribution), we find that the accuracy is generally very strong across models. We also evaluate on a fully out of distribution toy dataset (``simple adverbs") of the form {\tokens ``The traveller [adverb] walked to their destination. The traveller felt very"}. The results can be found in Figure \ref{fig:kmeans-accuracy-gpt2}. This is strongly suggestive that we are stumbling on a genuine representation of sentiment.

\subsubsection{Activation addition}
\begin{figure}[ht]
        \centering
        \includegraphics[width=\linewidth]{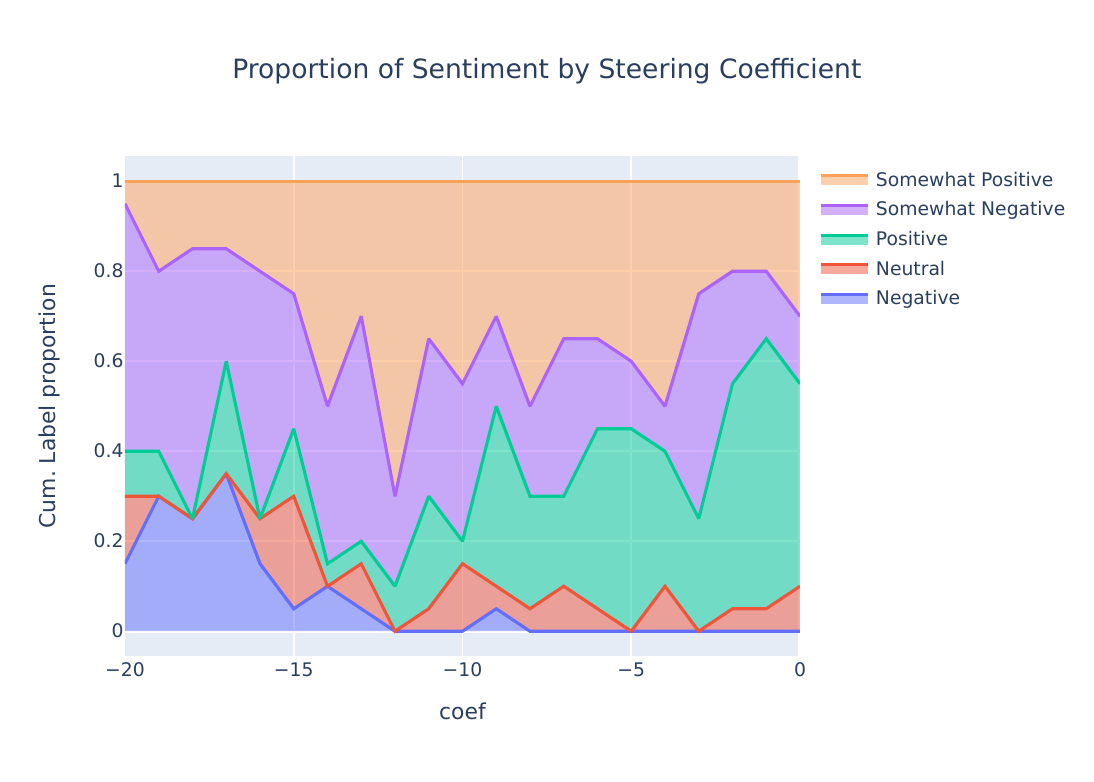}
        \caption{Area plot of sentiment labels for generated outputs by activation steering coefficient, starting from a single positive movie review continuation prompt. Activation addition \citep{turner2023activation} was performed in GPT2-small's first residual stream layer. Classification was performed by GPT-4.}
        \label{fig:steering-proportions}
    \end{figure}

We perform activation addition \citep{turner2023activation} on GPT2-small for a single positive simple movie review continuation prompt (from Section \ref{section:datasets}) in order to flip the generated outputs from negative to prompt. The ``steering coefficient" is the multiple of the sentiment direction which we add to the first layer residual stream. The outputs are extremely negative by the time we reach coefficient -17 and we observe a gradual transition for intermediate coefficients (Figure \ref{fig:steering-proportions}).

\subsubsection{Multi-lingual sentiment}
\begin{figure}[ht]

    \centering

    \begin{subfigure}[b]{\linewidth}
        \centering
        \includegraphics[width=\linewidth,clip,trim={0 14cm 0 0}]{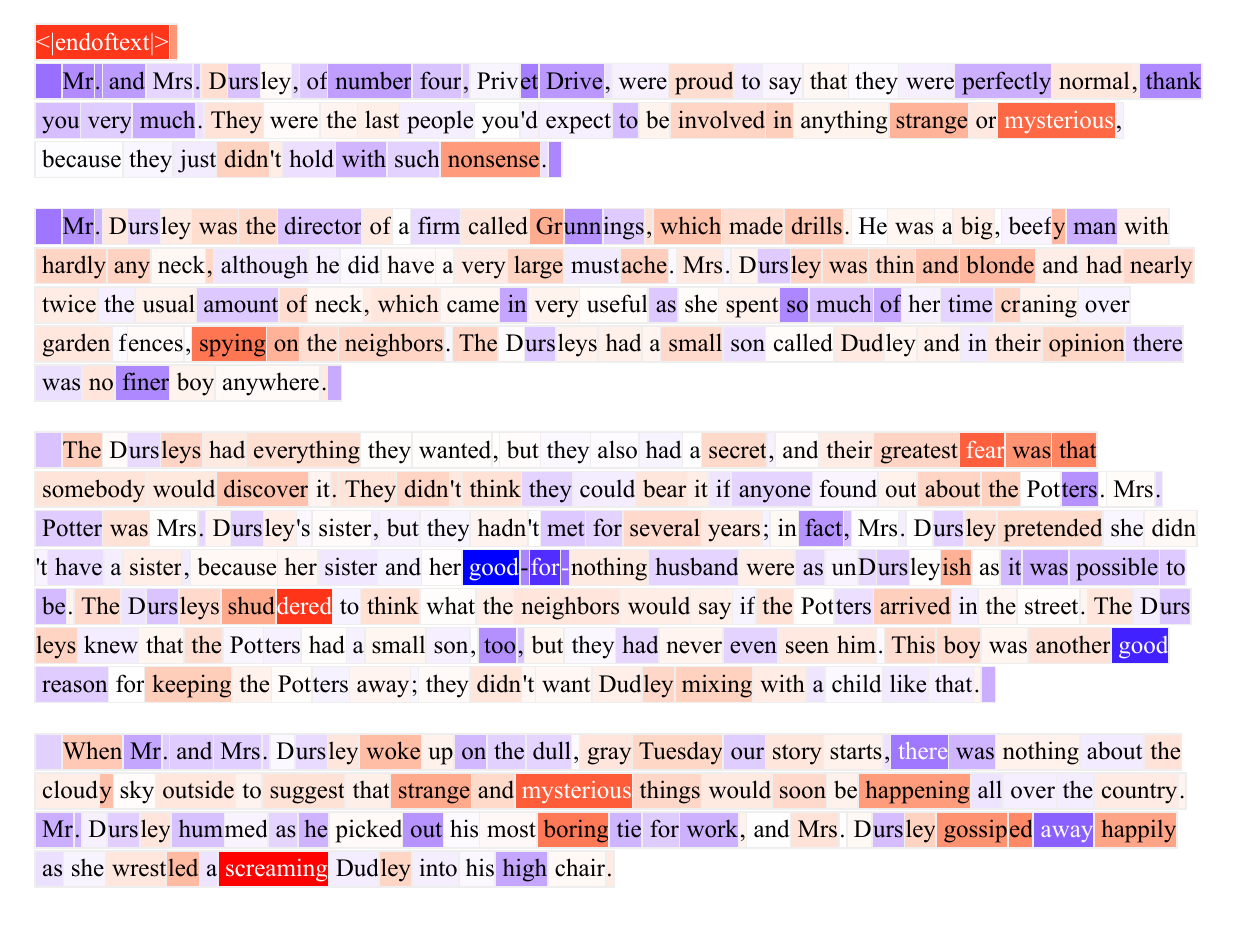}
        \caption{First 4 paragraphs of Harry Potter in English}
        \label{subfig:harry-potter-en}
    \end{subfigure}
    
    \begin{subfigure}[b]{\linewidth}
        \centering
        \includegraphics[width=\linewidth,clip,trim={0 11.8cm 0 0}]{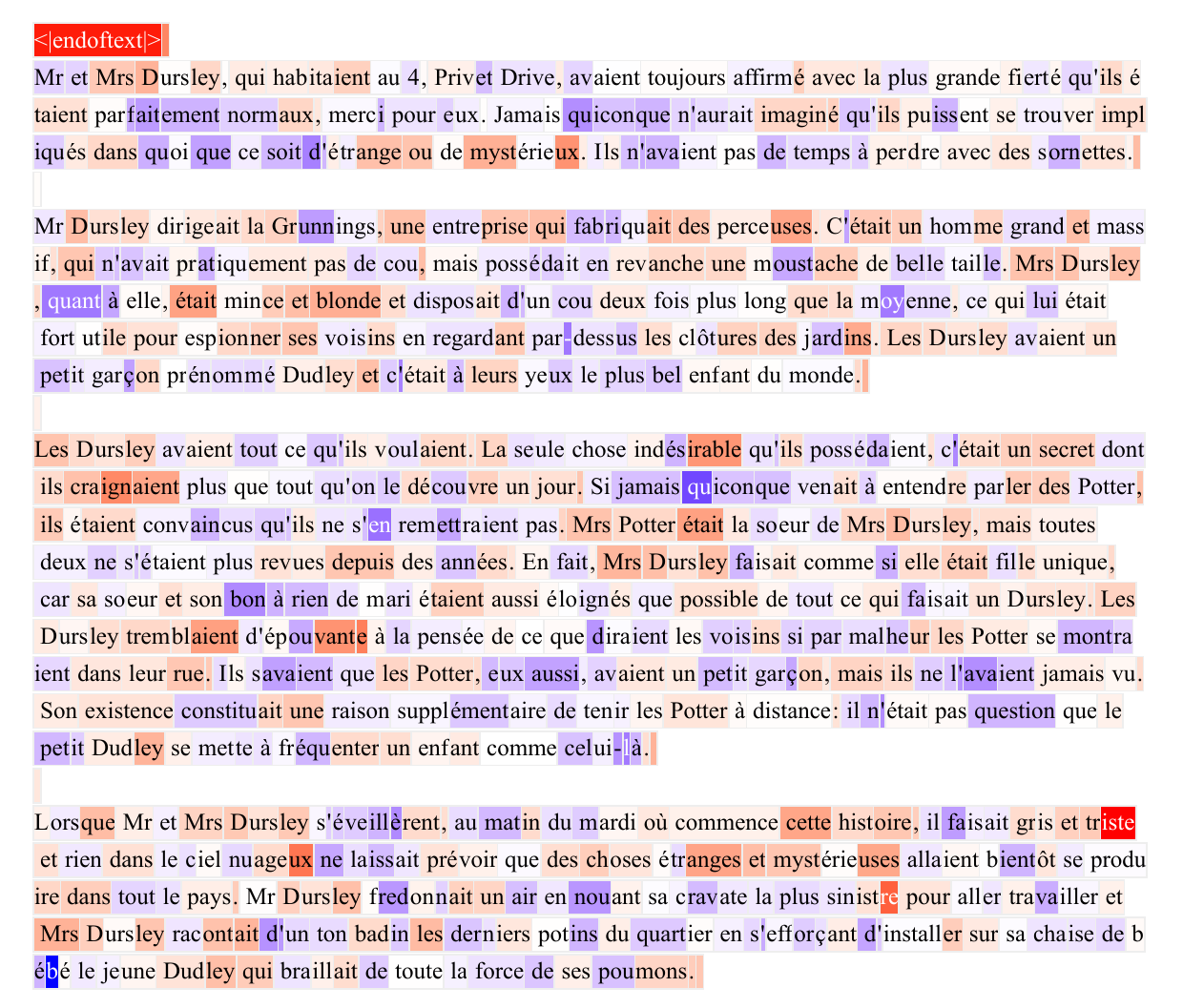}
        \caption{First 3 paragraphs of Harry Potter in French}
        \label{subfig:harry-potter-fr}
    \end{subfigure}

    \caption{First paragraphs of Harry Potter in different languages. Model: pythia-2.8b.}
    \label{fig:harry-potter}
\end{figure}

We use the first few paragraphs of Harry Potter in English and French as a standard text \citep{mathematicalframework}. We find that intermediate layers of pythia-2.8b demonstrate intuitive sentiment activations for the French text (Figure \ref{fig:harry-potter}). It is important to note that none of the models are very good at French, but this was the smallest model where we saw hints of generalisation to other languages. The representation was not evident in the first couple of layers, probably due to the poor tokenization of French words.

\subsubsection{Interpretability of negations}
\begin{figure}[ht]
    \centering
    \includegraphics[width=1\linewidth,clip,trim={0 1cm 4cm 0}]{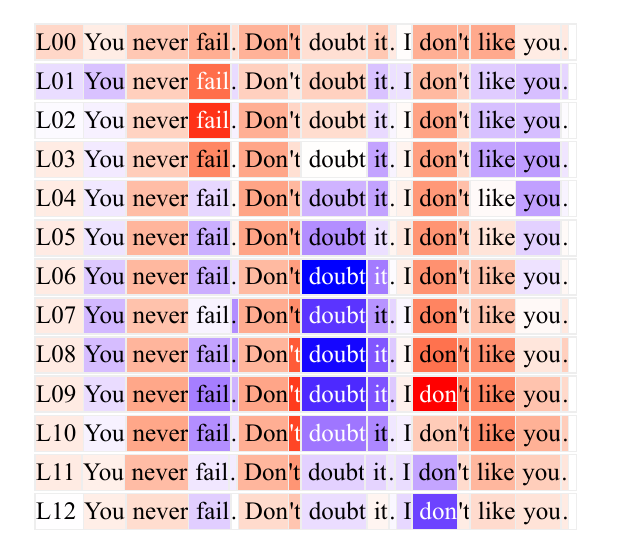}
    \caption{Visualizing the sentiment activations across layers for a text where the sentiment hinges on negations. Color represents sentiment activation at the given layer and position. Red is negative, blue is positive. Each row is a residual stream layer, first layer is at the top. The three sentences were input as a single prompt, but the pattern was extremely similar using separate prompts. Model: GPT2-small}
    \label{fig:negation}
\end{figure}

We visualise the sentiment activations for all 12 layers of GPT2-small simultaneously on the prompt {\tokens ``You never fail. Don't doubt it. I am not uncertain"} (Figure \ref{fig:negation}). This allows us to observe how {\tokens fail}, {\tokens doubt} and {\tokens uncertain} shift from negative to positive sentiment during the forward pass of the model.

\subsection{Is sentiment really a hyperplane?}\label{section:das-sweep}
\begin{figure}[ht]
        \centering
        \begin{subfigure}{0.5\linewidth}
        \centering
        \includegraphics[width=\linewidth,clip,trim={0 10cm 5cm 10cm}]{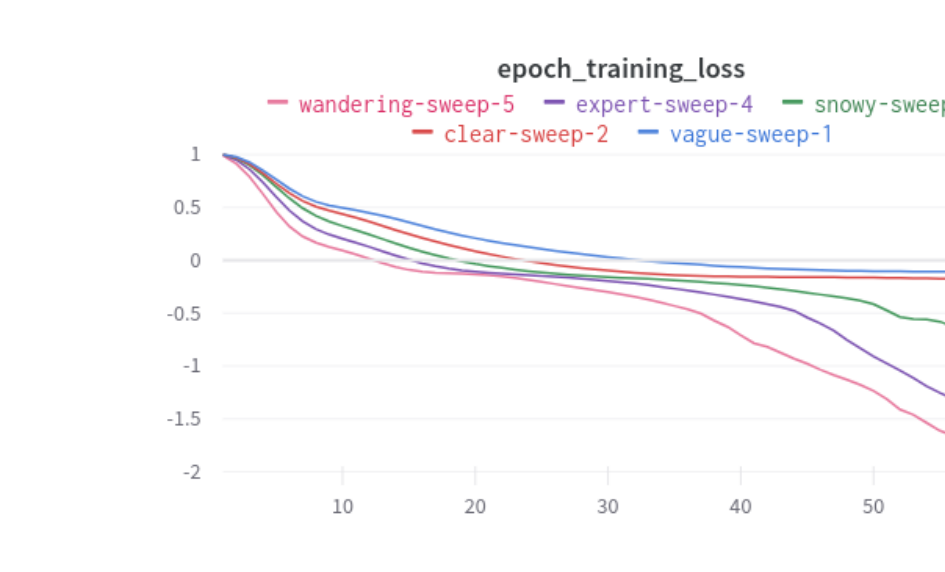}
        \caption{Training loss for DAS on adjectives in a toy movie review dataset}
        \label{fig:das-sweep-train-loss}
    \end{subfigure}

    \begin{subfigure}{0.5\linewidth}
        \centering
        \includegraphics[width=\linewidth,clip,trim={0 10cm 5cm 10cm}]{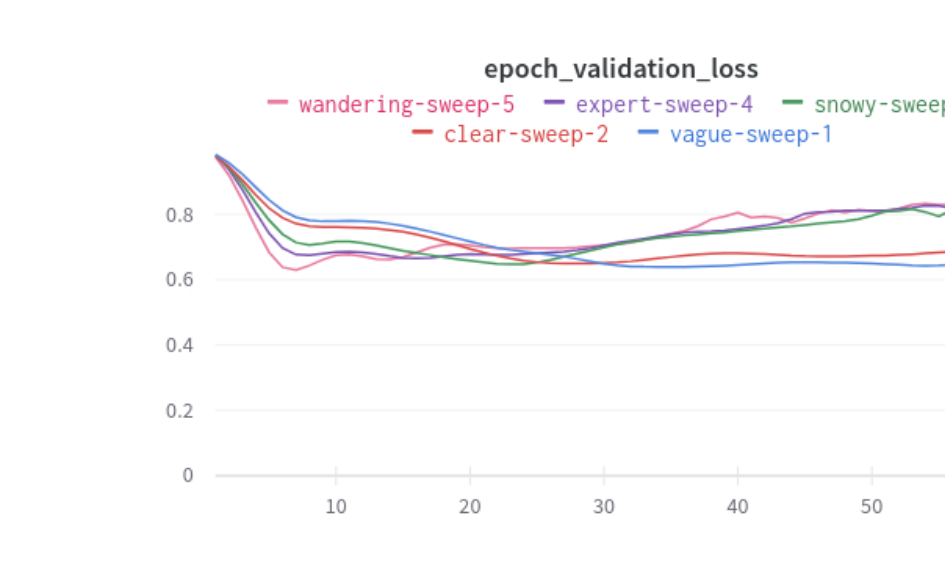}
        \caption{Validation loss for DAS on a simple character mood dataset with a varying adverb}
        \label{fig:das-sweep-val-loss}
    \end{subfigure}
        \caption{DAS sweep over the subspace dimension (GPT2-small). The runs are labelled with the integer $n$ where $d_\text{DAS}=2^{n-1}$. Loss is 1 minus the usual \gls{patching metric}.}
        \label{fig:das-sweep}
    \end{figure}

    In our \gls{directional patching} experiments, we have somewhat artificially selected just 1 dimension as our hypothesised structure for the sentiment subspace. We can perform DAS with any number of dimensions. Figure \ref{fig:das-sweep} demonstrates that whilst increasing the DAS dimension improves the \gls{patching metric} in-sample (\ref{fig:das-sweep-train-loss}), the metric does not improve out-of-distribution (\ref{fig:das-sweep-val-loss}).

    \subsection{Detailed circuit analysis}\label{section:detailed-circuit-analysis}
    In order to build a picture of each circuit, we used the process pioneered in \cite{ioi}:
    \begin{itemize}[leftmargin=0.5cm]
        \item Identify which model components have the greatest impact on the logit difference when path patching is applied (with the final result of the residual stream set as the receiver).
        \item Examine the attention patterns (value-weighted, in some cases) and other behaviors of these components (in practice, attention heads) in order to get a rough idea of what function they are performing.
        \item Perform path-patching using these heads (or a distinct cluster of them) as receivers.
        \item Repeat the process recursively, performing contextual analyses of each ``level" of attention heads in order to understand what they are doing, and continuing to trace the circuit backwards.
    \end{itemize}
    In each path-patching experiment, change in logit difference is used as the patching metric. We started with GPT-2 as an example of a classic LLM displays a wide range of behaviors of interest, and moved to larger models when necessary for the task we wanted to study (choosing, in each case, the smallest model that could do the task).

    \subsubsection{Simple sentiment - GPT-2 small}
\begin{figure}[h]
            \centering
            \includegraphics[width=\linewidth]{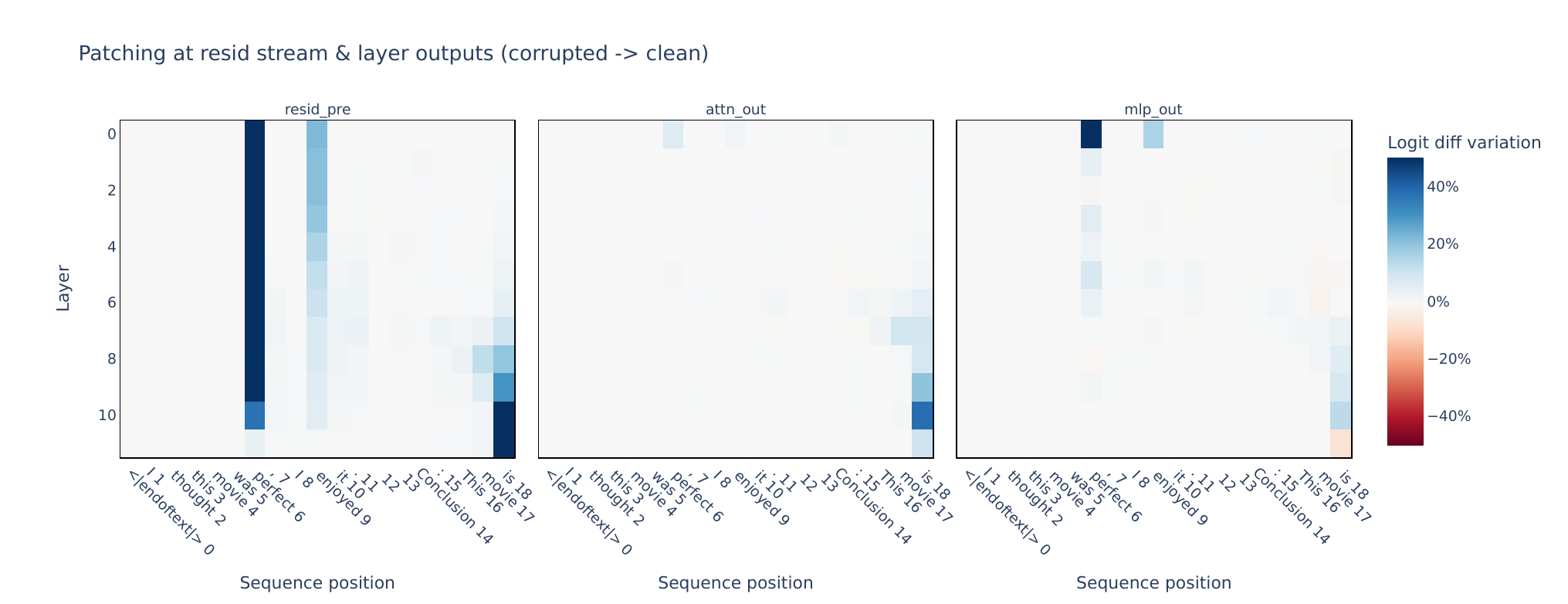}
            \caption{Activation patching results for the GPT-2 Small ToyMovieReview circuit, showing how much of the original logit difference is recaptured when swapping in activations from $x_{orig}$ (when the model is otherwise run on $x_{flipped}$). Note that attention output is only important at the {\tokens SUM} position, and that this information is important to task performance at the residual stream layers (8 and 9) in which the summary-readers reside. Other than this, the most important residual stream information lies at the ADJ and VRB positions.}
            \label{fig:toy-movie-review-circuit-patching}
        \end{figure}

    We examined the circuit performing tasks for the following sentence template:
    \[\text{\tokens I thought this movie was ADJECTIVE, I VERBed it. Conclusion: This movie is}\]
    Using a threshold of 5\%-or-greater damage to the logit difference for our patching experiments, we found that GPT-2 Small contained 4 primary heads contributing to the most proximate level of circuit function--10.4, 9.2, 10.1, and 8.5 (using ``layer.head" notation). Examining their value-weighted attention patterns, we found that attention to {\tokens ADJ} and {\tokens VRB} in the sentence was most prominent in the first three heads, but 8.5 attended primarily to the second ``{\tokens movie}" token. We also observed that 9.2 attended to this token as well as to {\tokens ADJ}. (Results of activation patching can be seen in Fig. \ref{fig:toy-movie-review-circuit-patching}.)

    Conducting path-patching with 8.5 and 9.2 as receivers, we identified two heads--7.1 and 7.5--that primarily attend to {\tokens ADJ} and {\tokens VRB} from the ``{\tokens movie}" token. We further determined that the output of these heads, when path-patched through 9.2 and 8.5 as receivers, was causally important to the circuit (with patching causing a logit difference shift of 7\% and 4\% respectively for 7.1 and 7.5). This was not the case for other token positions, which demonstrates that causally relevant information is indeed being specially written to the ``{\tokens movie}" position. We thus designated it the {\tokens SUM} token in this circuit, and we label 8.5 a summary-reader head.

    Repeating our analysis with lower thresholds yielded more heads with the same behavior but weaker effect sizes, adding 9.10, 11.9, and 6.4 as summary reader, direct sentiment reader, and sentiment summarizer respectively. This gives a total of 9 heads making up the circuit. 

    \subsubsection{Multi-subject mood stories circuit - Pythia 2.8b}
    We also examined the circuit for this sentence template:
    {\tokens Carl hates parties, and avoids them whenever possible. Jack loves parties, and joins them whenever possible. One day, they were invited to a grand gala. Jack feels very} [{\tokens excited}/{\tokens nervous}].
    We did not attempt to reverse-engineer the entire circuit, but examined it from the perspective of what matters causally for sentiment processing--especially determining to what extent summarization occurred.

    Following the same process as with GPT-2 with preference/sentiment-flipped prompts (that is, taking $x_{orig}$ to be ``{\tokens John hates parties,... Mary loves parties}," and $x_{flipped}$ to be ``{\tokens John loves parties,... Mary hates parties}"), we initially identified 5 key heads that were most causally important to the logit difference at {\tokens END}: 17.19, 22.5, 14.4, 20.10, and 12.2 (in ``layer.head" notation). Examining the value-weighted attention patterns, we observed that the top token receiving attention from {\tokens END} was always the repeated name {\tokens RNAME} (e.g., ``{\tokens John}" in ``{\tokens John feels very}") or the ``{\tokens feels}" token {\tokens FEEL}, indicating that some summarization may have taken place there.

    We also observed that the top token attended to from {\tokens RNAME} and {\tokens FEEL} was in fact the comma at the end of the queried preference phrase (that is, the comma at the end of ``{\tokens John hates parties}"). We designate this position {\tokens COMMASUM}.

    \paragraph{Multi-functional heads} Interestingly, we observed that most of these heads were multi-functional: that is, they both attended to {\tokens COMMASUM} from {\tokens RNAME} and {\tokens FEEL}, and also attended to {\tokens RNAME} and {\tokens FEEL} from {\tokens END}, producing output in the direction of the logit difference. This is possible because these heads exist at different layers, and later heads can read the summarized information from previous heads as well as writing their own summary information. 
    
    \paragraph{Direct effect heads} Specifically, the direct effect heads were:
    \begin{itemize}
        \item Head 17.19 did not attend to commas significantly, but did attend to the periods at the end of each preference sentence in addition to its primary attention to {\tokens RNAME} and {\tokens FEEL}, and did not display {\tokens COMMASUM}-reading behavior.
        \item Head 22.5 attended almost exclusively to {\tokens FEEL}, and did not display {\tokens COMMASUM}-reading behavior.
        \item Other direct effect heads (14.4, 20.10 and 12.2) did show {\tokens COMMASUM}-reading behavior as well as reading from the near-end tokens to produce output in the direction of the logit difference. In each case, we verified with path-patching that information from these positions was causally relevant.
    \end{itemize}

    \paragraph{Name summary writers} We also found important heads (12.17 being by far the most important) that are only engaged with attending to {\tokens COMMASUM} and producing output at {\tokens RNAME} and {\tokens FEEL}.

    \paragraph{Comma summary writers} We further investigated what circuitry was causally important to task performance mediated through the {\tokens COMMASUM} positions, but did not flesh this out in full detail; after finding initial examples of summarization, we focused on its causal relevance and interaction with the sentiment direction, leaving deeper investigation to future work.

    \subsection{Additional summarization findings}
        \paragraph{Circuitry for processing commas vs. original phrases is semi-separate}
                Though there is overlap between the attention heads involved in the circuitry for processing sentiment from key phrases and that from summarization points, there are also some clear differences, suggesting that the ability to read summaries could be a specific capability developed by the model (rather than the model simply attending to high-sentiment tokens).
    
                \begin{figure}
                    \centering
                    \includegraphics[scale=0.4]{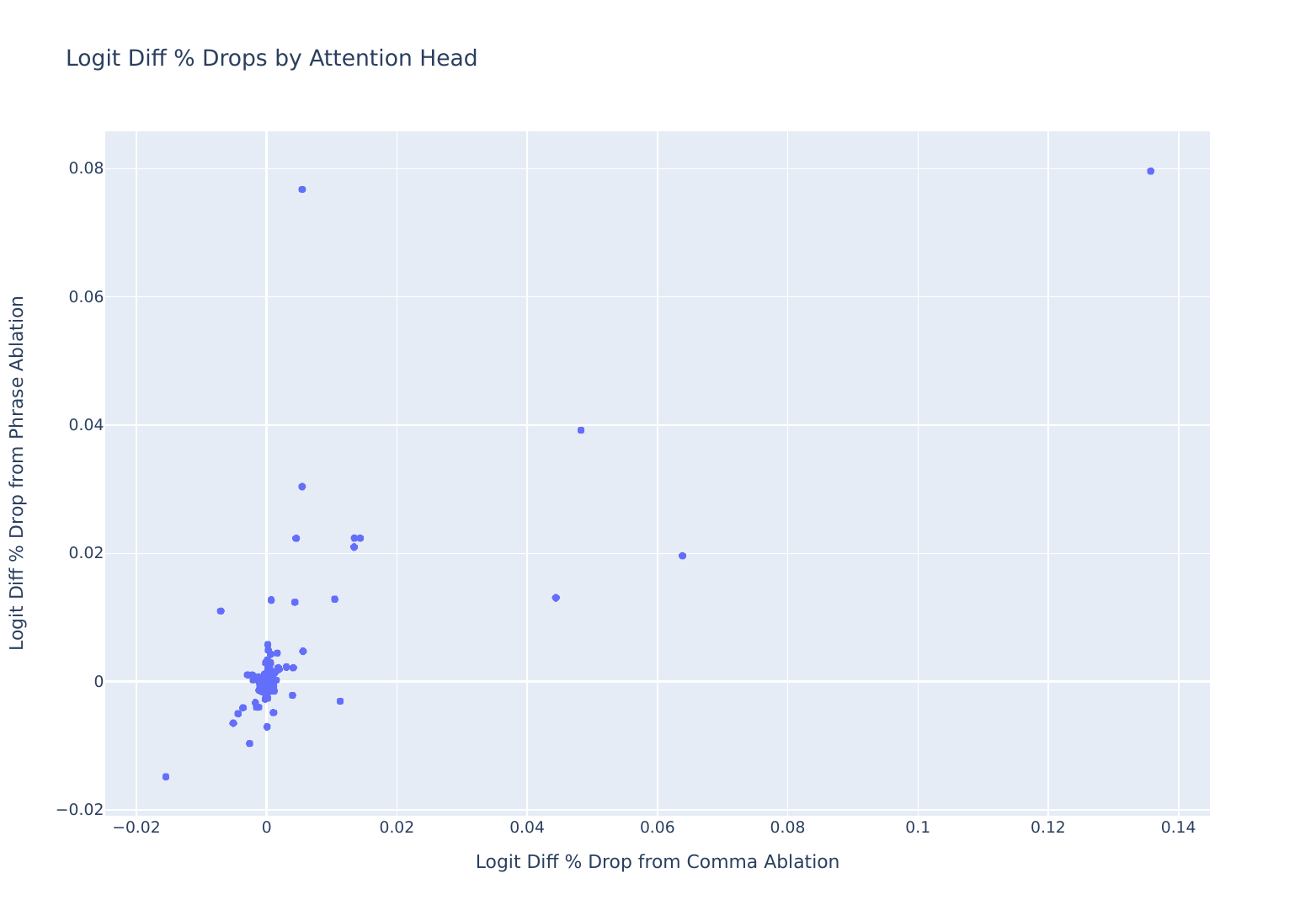}
                    \caption{Logit difference drops by head when commas or pre-comma phrases are patched. Model: pythia-2.8b.}
                    \label{fig:comma_vs_phrase_ld_drops_by_head}
                \end{figure}
    
                As can be seen in Figure \ref{fig:comma_vs_phrase_ld_drops_by_head}, there are distinct groups of attention heads that result in damage to the logit difference in different situations--that is, some react when phrases are patched, some react disproportionately to comma patching, and one head seems to have a strong response for either patching case. This is suggestive of semi-separate summary-reading circuitry, and we hope future work will result in further insights in this direction.

                \begin{figure}[h]
                    \centering
                    \includegraphics[width=1.0\linewidth]{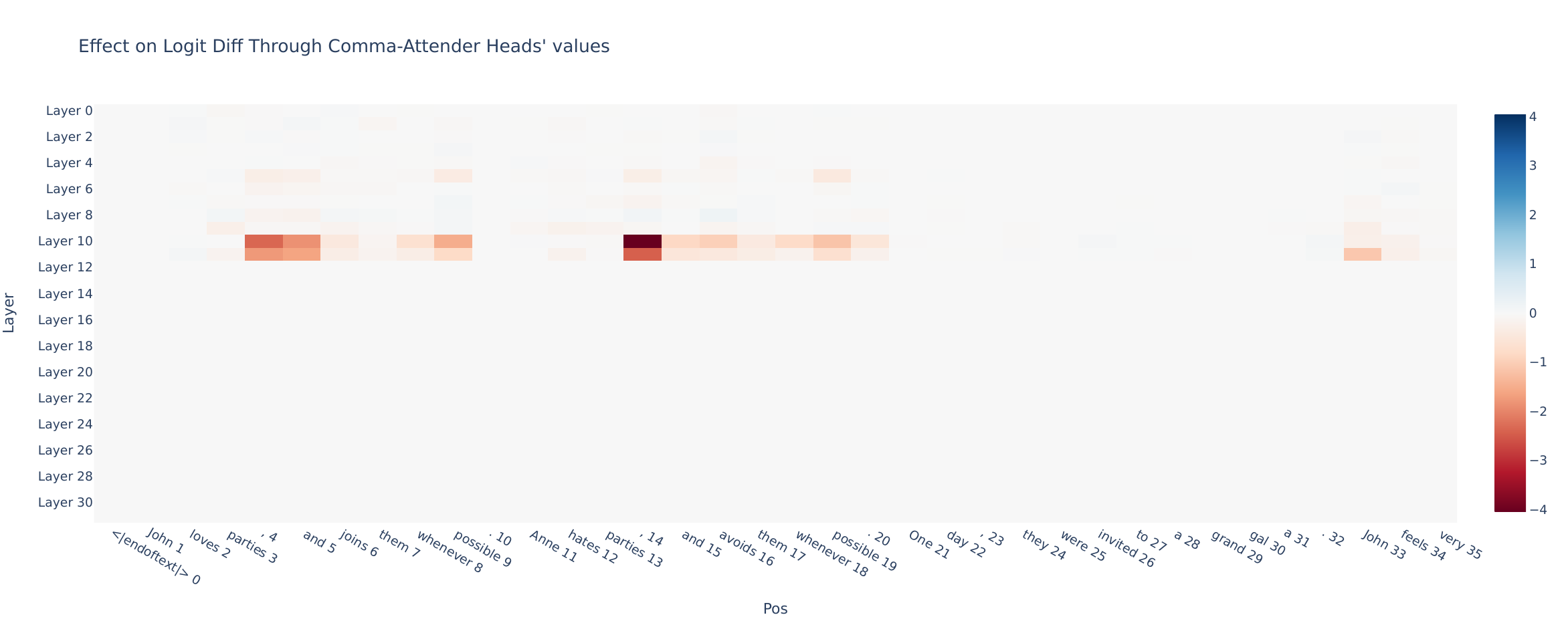} %
                    \caption{Path-patching commas and comma phrases in Pythia-2.8b, with attention heads L12H2 and L12H17 writing to repeated name and "feels" as receivers. Patching the paths between the comma positions and the receiver heads results in the greatest performance drop for these heads.}
                    \label{fig:pythia-2.8b-path_patching_commas}
                \end{figure}

    \subsection{Neurons writing to sentiment direction in GPT2-small are interpretable}\label{section:interpreting-neurons}
    \begin{figure}[ht]
        \centering
        \includegraphics[width=1\linewidth]{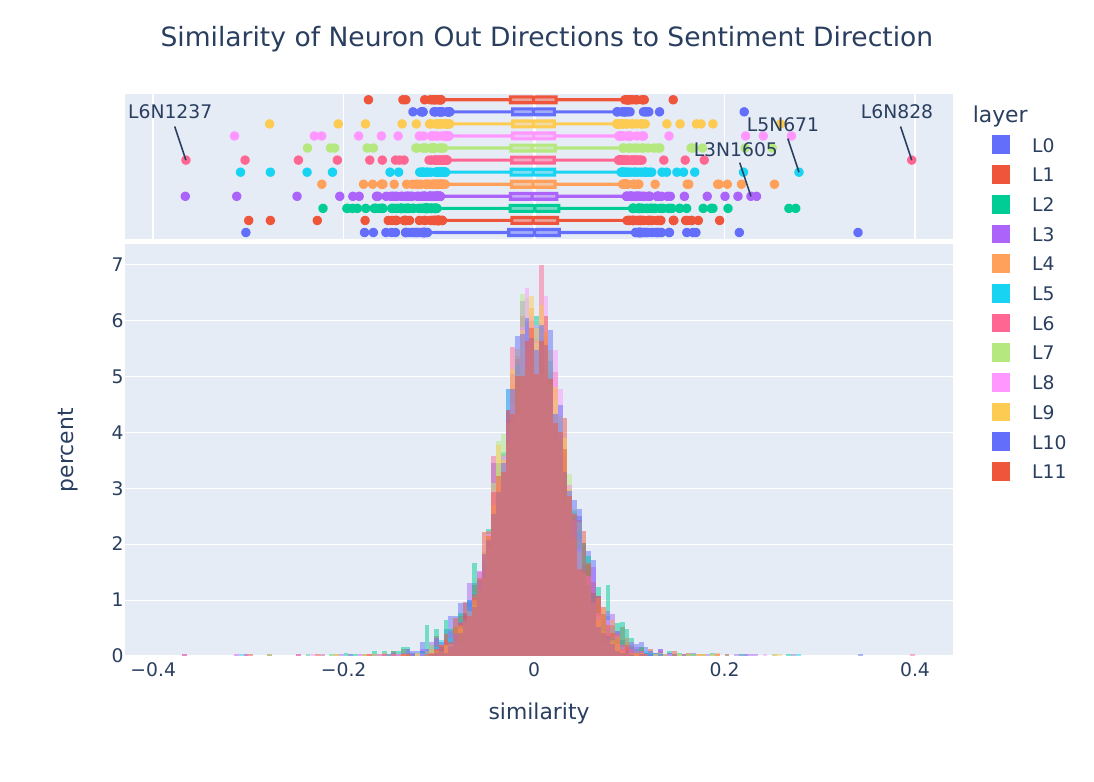}
        \caption{Cosine similarity of neuron out-directions and the sentiment direction in GPT2-small}
        \label{fig:out-similarity}
    \end{figure}
    We observed that the cosine similarities of neuron out-directions with the sentiment direction are extremely heavy tailed (Figure \ref{fig:out-similarity}). Thanks to Neuroscope \citep{nanda_neuroscope_io}, we can quickly see whether these neurons are interpretable. Indeed, here are a few examples from the tails of that distribution:
    \begin{itemize}
        \item \href{https://neuroscope.io/gpt2-small/3/1605.html}{L3N1605} activates on {\tokens ``hesitate"} following a negation
        \item Neuron \href{https://neuroscope.io/gpt2-small/6/828.html}{L6N828} seems to be activating on words like {\tokens ``however"} or {\tokens ``on the other hand"} \textit{if} they follow something negative
        \item Neuron \href{https://neuroscope.io/gpt2-small/5/671.html}{L5N671} activates on negative words that follow a {\tokens ``not"} contraction (e.g. {\tokens didn't}, {\tokens doesn't}) 
        \item \href{https://neuroscope.io/gpt2-small/6/1237.html}{L6N1237} activates strongly on {\tokens ``but"} following {\tokens ``not bad"}
    \end{itemize}

    We take L3N1605, the {\tokens ``not hesitate"} neuron, as an extended example and trace backwards through the network using Direct Logit Attribution\footnote{This technique decomposes model outputs into the sum of contributions of each component, using the insight from \citet{mathematicalframework} that components are independent and additive}. We computed the relative effect of different model components on L3N1605 in the two different cases {\tokens ``I would not hesitate"} vs. {\tokens ``I would always hesitate"}. The main contributors to this difference are L1H0, L3H10, L3H11 and MLP2. Expanding out MLP2 into individual neurons we find that the contributions to L3N1605 are sparse. For example, L2N1154 activates on words like {\tokens ``don't"}, {\tokens ``not"}, {\tokens ``no"}, etc. It activates on {\tokens ``not"} but not {\tokens ``hesitate"} in {\tokens ``I would not hesitate"} but activates on {\tokens ``hesitate"} in {\tokens ``I would always hesitate"}. Visualizing the attention pattern of L1H0 shows that it attends from {\tokens ``hesitate"} to the previous token if it is {\tokens ``not"}, but not if it is {\tokens ``always"}.

    These anecdotal examples suggest at a complex network of machinery for transmitting sentiment information across components of the network using a single critical axis of the residual stream as a communication channel. We think that exploring these neurons further could be a very interesting avenue of future research, particularly for understanding how the model updates sentiment based on negations where these neurons seem to play a critical role.

        \subsection{Detailed description of metrics}\label{section:detailed-metrics}
            \begin{itemize}
                \item \textbf{Logit Difference:} We extend the logit difference metric used by \citet{ioi} to the setting with 2 \textit{classes} of next token rather than only 2 valid next tokens. This is useful in situations where there are many possible choices of positively or negatively valenced next tokens. Specifically, we examine the average difference in logits between sets of positive/negative next-tokens $T^\text{positive}=\{t^\text{positive}_i: 1 \leq i \leq n\}$ and $T^\text{negative}=\{t_i^\text{negative}: 1 \leq i \leq n\}$ in order to get a smooth measure of the model's ability to differentiate between sentiment. That is, we define the logit difference as $\frac{1}{n}\sum_i \left[\mathrm{logit}(t^\text{positive}_i) - \mathrm{logit}(t_i^\text{negative})\right]$. Larger differences indicate more robust separation of the positive/negative tokens, and zero or inverted differences indicate zero or inverted sentiment processing respectively. When used as a \gls{patching metric}, this demonstrates the causal efficacy of various interventions like activation patching or ablation.\footnote{We use this metric often because it is more sensitive than accuracy to small shifts in model behavior, which is particularly useful for circuit identification where the effect size is small but real. That is, in many cases a token of interest might become much more likely but not cross the threshold to change accuracy metrics, and in this case logit difference will detect it. Logit difference is also useful when trying to measure the model behavior transition between two different, opposing prompts--in this case, the logit difference for each of the prompts is used for lower and upper baselines, and we can measure the degree to which the logit difference behavior moves from one pole to the other.}
                \item \textbf{Logit Flip:} Similar to logit difference, this is the percentage of cases where the logit difference between $T^\text{positive}$ and $T^\text{negative}$ is inverted after a causal intervention. This is a more discrete measure which is helpful for gauging whether the magnitude of the logit differences is sufficient to actually flip model predictions.
                \item \textbf{Accuracy:} Out of a set of prompts, the percentage for which the logits for tokens $T^\text{correct}$ are greater than $T^\text{incorrect}$. In practice, usually each of these sets only has one member (e.g., ``Positive" and ``Negative").
            \end{itemize}

\subsection{Toy dataset details}\label{section:toy-movie-review-details}
The ToyMovieReview dataset consists of prompts of the form {\tokens "I thought this movie was ADJ, I VRB it. [NEWLINE] Conclusion: This movie is"}. We substituted different adjective and verb tokens into the two variable placeholders to create a prompt for each distinct adjective. We averaged the logit difference across 5 positive and 5 negative completions to determine whether the continuation was positive or negative.

\begin{verbatim}
positive_adjectives_train:
  - perfect
  - fantastic
  - delightful
  - cheerful
  - good
  - remarkable
  - satisfactory
  - wonderful
  - nice
  - fabulous
  - outstanding
  - satisfying
  - awesome
  - exceptional
  - adequate
  - incredible
  - extraordinary
  - amazing
  - decent
  - lovely
  - brilliant
  - charming
  - terrific
  - superb
  - spectacular
  - great
  - splendid
  - beautiful
  - positive
  - excellent
  - pleasant

negative_adjectives_train:
  - dreadful
  - bad
  - dull
  - depressing
  - miserable
  - tragic
  - nasty
  - inferior
  - horrific
  - terrible
  - ugly
  - disgusting
  - disastrous
  - annoying
  - boring
  - offensive
  - frustrating
  - wretched
  - inadequate
  - dire
  - unpleasant
  - horrible
  - disappointing
  - awful

positive_adjectives_test:
  - stunning
  - impressive
  - admirable
  - phenomenal
  - radiant
  - glorious
  - magical
  - pleasing
  - lively
  - warm
  - strong
  - helpful
  - vivid
  - modern
  - crisp
  - sweet

negative_adjectives_test:
  - foul
  - vile
  - appalling
  - rotten
  - grim
  - dismal
  - lazy
  - poor
  - rough
  - noisy
  - sour
  - flat
  - ancient
  - bitter

positive_verbs:
  - enjoyed
  - loved
  - liked
  - appreciated
  - admired

negative_verbs:
  - hated
  - disliked
  - despised

positive_answer_tokens:
  - great
  - amazing
  - awesome
  - good
  - perfect

negative_answer_tokens:
  - terrible
  - awful
  - bad
  - horrible
  - disgusting
\end{verbatim}

The ToyMoodStories dataset consists of prompts of the form ``{\tokens NAME1 VRB1.1 parties, and VRB1.2 them whenever possible. NAME2 VRB2.1 parties, and VRB2.2 them whenever possible. One day, they were invited to a grand gala. QUERYNAME feels very}". To evaluate the model's output, we measure the logit difference between the ``{\tokens excited}" and ``{\tokens nervous}" tokens.

{\tokens VRB1.1} and {\tokens VRB2.1} are always one of: \begin{verbatim}
    - hates
    - loves
\end{verbatim} and {\tokens VRB1.2} and {\tokens VRB2.2} are always one of: \begin{verbatim}
    - avoids
    - joins
\end{verbatim}

In each case, the two verbs in each sentence will agree in sentiment, and the sentence with {\tokens NAME1} will always have opposite sentiment to that of {\tokens NAME2}.

Names are sampled from the following list:
\begin{verbatim}
    - John
    - Anne
    - Mark
    - Mary
    - Peter
    - Paul
    - James
    - Sarah
    - Mike
    - Tom
    - Carl
    - Sam
    - Jack
\end{verbatim}

Each combination of {\tokens NAME1, NAME2, QUERYNAME} are included in the dataset (where half the time {\tokens QUERYNAME} matches the first name, and half the time it matches the second). Where necessary for computational tractability, we take a subsample of the first 16 items of this dataset.

    \subsection{Glossary}
    \label{section:glossary}

    \printglossaries

\end{document}